\documentclass[smallextended]{svjour3}       
\usepackage[T1]{fontenc}
\usepackage{graphicx}
\usepackage{amsmath}
\usepackage{amssymb}
\usepackage[vlined,boxed,ruled]{algorithm2e}
\usepackage{array}
\usepackage{url}
\usepackage{float}
\usepackage{xcolor}
\usepackage{hyperref}
\usepackage{pdflscape}
\usepackage{subfigure}
\usepackage{booktabs}
\usepackage{siunitx}
\usepackage{tabularx}
\usepackage[toc,page]{appendix}

\let\originalleft\left
\let\originalright\right
\renewcommand{\left}{\mathopen{}\mathclose\bgroup\originalleft}
\renewcommand{\right}{\aftergroup\egroup\originalright}

\definecolor{citecol}{rgb}{0,0,0.5}
\hypersetup{urlcolor=citecol,linkcolor=black,citecolor=citecol,colorlinks=true}

\newcommand{\ie}{\textit{i.e.}, }

\newcommand{\modif}[1]{{\color{black}#1}}
\newcommand{\modiff}[1]{{\color{black}#1}}

\DeclareMathOperator*{\argmin}{arg\,min}
\DeclareMathOperator*{\argmax}{arg\,max}

\newcommand{\ourRepo}{\url{https://github.com/fpetitjean/ProximityForest/}}

\begin{document}
\sloppy

\title{Proximity Forest\\ An effective and scalable distance-based classifier for time series}

\titlerunning{Proximity Forest}        

\author{Benjamin~Lucas \and Ahmed Shifaz \and Charlotte~Pelletier \and Lachlan~O'Neill \and Nayyar~Zaidi \and   Bart~Goethals  \and Fran\c{c}ois~Petitjean \and Geoffrey~I.~Webb}

 \institute{
               Faculty of Information Technology\\
               25 Exhibition Walk\\
               Monash University, Melbourne \\
               VIC 3800, Australia\\
              \email{\{benjamin.lucas,charlotte.pelletier,ahmed.shifaz,nayaar.zaidi,lsone1\}@monash.edu}\\
              \email{\{bart.goethals,francois.petitjean,geoff.webb\}@monash.edu}
 }

\maketitle

\begin{abstract}
Research into the classification of time series has made enormous progress in the last decade. 
The UCR time series archive has played a significant role in challenging and guiding the development of new learners for time series classification. 
The largest dataset in the UCR archive holds 10 \emph{thousand} time series only; which may explain why the primary research focus has been on creating algorithms that have high accuracy on relatively small datasets. 


This paper introduces Proximity Forest, an algorithm that learns accurate models from datasets with millions of time series, and classifies a time series in milliseconds. 
\modif{The models are ensembles of highly randomized Proximity Trees.
Whereas conventional decision trees branch on attribute values (and usually perform poorly on time series), Proximity Trees branch on the proximity of time series to one exemplar time series or another; allowing us to leverage the decades of work into developing relevant measures for time series.
Proximity Forest gains both efficiency and accuracy by stochastic selection of both exemplars and similarity measures.}

Our work is motivated by recent time series applications that provide orders of magnitude more time series than the UCR benchmarks. Our experiments demonstrate that Proximity Forest is highly  competitive on the UCR archive: it ranks among the most accurate classifiers while being significantly faster. We demonstrate on a 1M time series Earth observation dataset that Proximity Forest retains this accuracy on datasets that are many orders of magnitude greater than those in the UCR repository, while learning its models at least 100,000 times faster than current state of the art models Elastic Ensemble and COTE.

\end{abstract}

\section{Introduction}
\label{sec:Introduction}
\modif{A growing number of time series applications address training from orders of magnitude more series than the largest in the benchmark UCR repository--- the 8,926 training series ElectricDevices data set. In contrast, the phoneme dataset \cite{hamooni2014phoneme} contains 370,000 series. The satellite dataset \cite{tan2017indexing} contains 1,000,000 series.}
The prior state-of-the-art in time series classification does not scale 
to such quantities.
In 2017, a meticulous study was conducted to compare the behaviour of the state-of-the-art \cite{bagnall2017great}. The authors draw the following conclusions: 
\begin{enumerate}
\item  The state-of-the-art is led by four classifiers that are: Collection of Transformation Ensembles (COTE) \cite{bagnall2015cote}, Elastic Ensembles (EE) \cite{lines2015ee}, Shapelet Transform (ST) \cite{hills2014shapelets}  and Bag of SFA Symbols (BOSS) \cite{schafer2015boss}. 
\item COTE is a special case in that it subsumes two of the other classifiers: it is a large ensemble classifier that includes EE and ST as sub-classifiers; COTE is on average, ``clearly superior to other published techniques.''
\item COTE's runtime complexity is bounded by (a) Shapelet Transform, which is $O(n^2\cdot l^4)$ \cite{bagnall2015cote} for $n$ time series of length $l$, and (b) the parameter searches for EE, some of which are $O(n^2\cdot l^3)$. The authors conclude ``An algorithm that is faster than COTE but not significantly less accurate would be a genuine advance in the field.''
\end{enumerate}
This is the challenge we tackle in this paper: developing an algorithm that is competitive with the accuracy of the state-of-the-art, but can learn from datasets with millions of time series. 
We call our algorithm \emph{Proximity Forest}. It is a tree based ensemble that makes the most of the decades of research into developing consistent similarity measures for time series. 
\begin{figure}
\centering
\subfigure[]{\includegraphics[width=.48\linewidth]{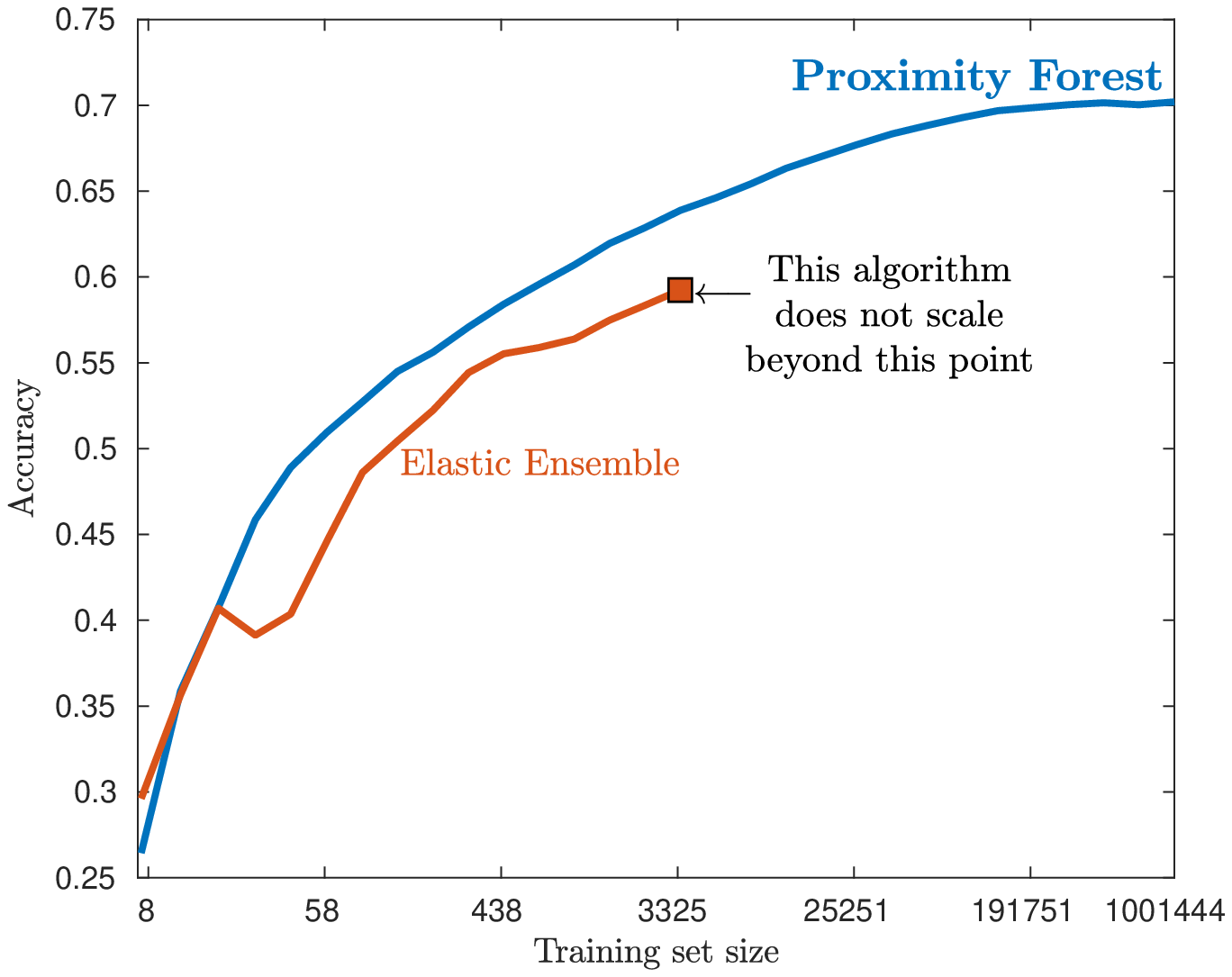}}
\subfigure[]{\includegraphics[width=.48\linewidth]{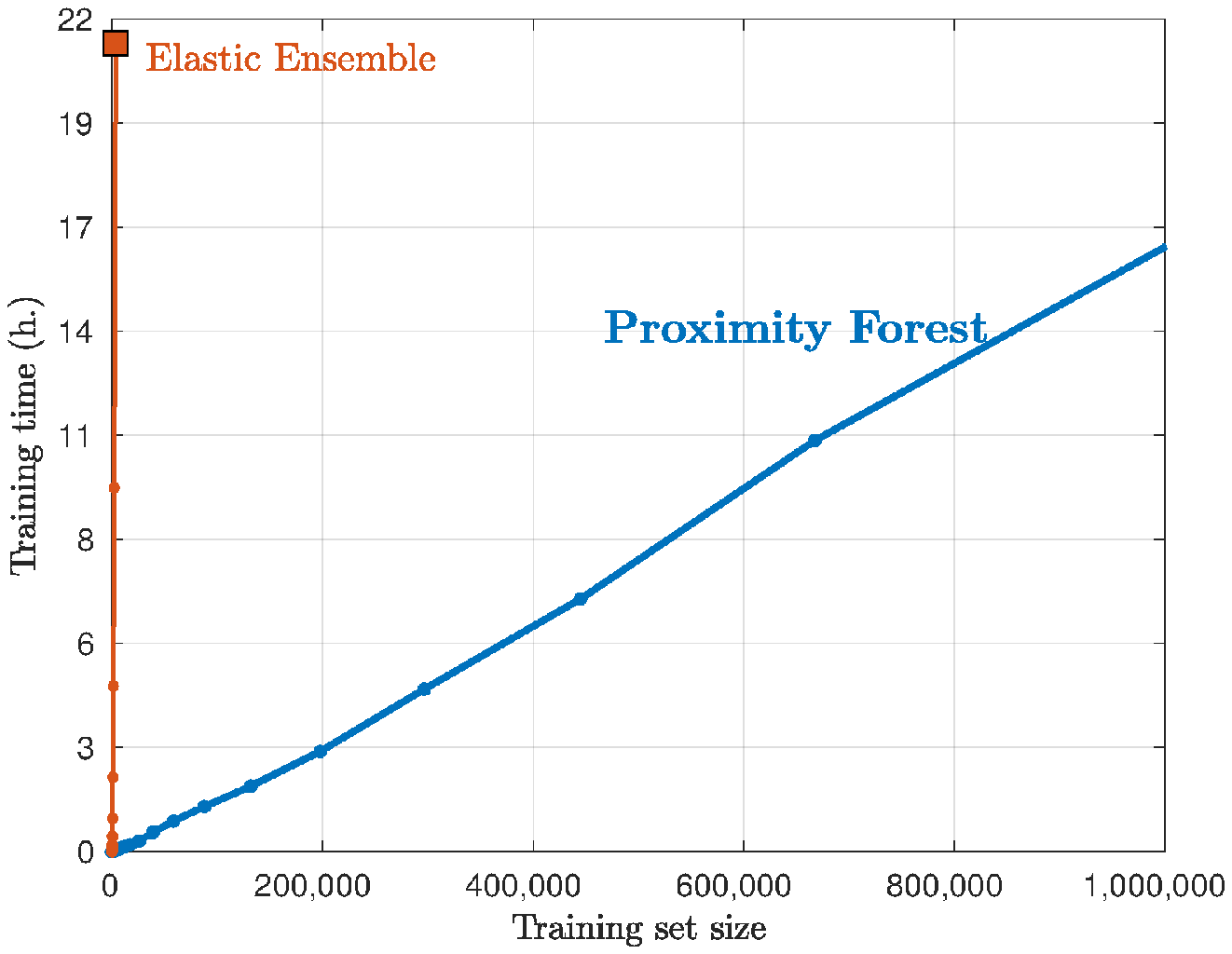}}
\caption{\label{fig:intro-train-time}Comparison of Proximity Forest (in blue) with Elastic Ensemble (in red). (a) Classification accuracy as a function of the size of the training data. (b) Training time as a function of the size of the training data. Note that we take Elastic Ensemble because it is the classifier that prevents scalability of the state-of-the-art COTE ensemble, which includes EE as one of its classifiers. }
\end{figure}


Typical decision trees branch on the value of an attribute. Treating the values at each time stamp as belonging to a single attribute does not work well on time series because the relevant signals are not necessarily aligned by time stamp.
Instead, Proximity Forest branches on the \emph{proximity} of a query time series to a set of reference series.
`Proximity' is defined by a given (time series) similarity measure and a set of parameters (most time series measures have parameters that are critical to their proper function). 
Our trees define separating hyperplanes for which the position is supported by time series themselves (whereas a traditional tree would split using a threshold on the value of an attribute). 
Proximity Forest, as opposed to nearest neighbour approaches, truly abstracts a model from data, which makes it possible to classify with time that is logarithmic with respect to training set size, as opposed to linear time for EE and COTE. 
Moreover, we will show that we specifically designed its training to scale linearly with the quantity of data, as opposed to at least quadratically for EE, COTE and ST.


\modiff{
Figure~\ref{fig:intro-train-time} shows the accuracy and the training time required by Proximity Forest and EE on our satellite dataset with increasing training set size. Two important elements are illustrated: (1) Proximity Forest scales linearly with training set size while EE scales quadratically; and, (2) Proximity Forest's classification accuracy on this dataset is substantially better than EE, even when they train on the same quantity of data. 
For our application, Proximity Forest can learn from 1M time series in 17 hours (on 1 CPU) while it would take over 200 years for EE --~a 103,000x speedup. Furthermore, ST and COTE learn slower than EE and thus will have \emph{even larger} training times. Note that EE is a component of COTE and hence sets a lower bound on COTE's training time. Our discussion will often focus on EE because EE is very similar to our algorithm in that it is trying to leverage existing time series similarity measures.

The virtues of Proximity Forest are not limited to large datasets, however. Our experiments show that it also outperforms EE in classification accuracy on the majority of the datasets of the UCR Archive.}

The remainder of this paper is structured as follows: in Section~\ref{sec:Related work}, we review the state-of-the-art in time series classification, with a particular focus on scalability. 
We then introduce Proximity Forest in Section~\ref{sec:Proximity Forest}. 
Our experiments (Section~\ref{sec:Experiments}) show that Proximity Forest (1) outperforms all other scalable algorithms on our case study in both accuracy and training time; and (2) is competitive with the state-of-the-art on UCR data in terms of accuracy. The section ends with a study of Proximity Forest's parameters and a discussion of how their values vary the results.

\section{Time series classification -- related work}
\label{sec:Related work}
We present here a non-exhaustive review of the state-of-the-art in time series classification and similar decision tree-based algorithms. We focus on our particular interest in this paper: scalable training and classification.

\subsection{Distance-based classification}
\subsubsection{Distances}
Time series have particular properties that have led to the development of specific similarity measures: they are often auto-correlated (the value of the time series at a timestamp is likely to be close to the ones just before and after), and often include non-linear distortions in the time axis (for example, because the start of the phenomenon of interest is delayed, or because sections of the phenomena are faster or slower). 
This has rendered typical similarity measures severely flawed and led to the development of specific similarities, of which most have an ability to re-align the series along a common intrinsic time-line.
Important measures
include Dynamic Time Warping (DTW) \cite{sakoe1971dynamic,sakoe1978dynamic}, Derivative DTW  (DDTW) \cite{keogh2001derivative,gorecki2013using}, Weighted DTW (WDTW) \cite{jeong2011weighted}, Longest Common Subsequence (LCSS) \cite{vlachos2006indexing}, Edit Distance with Real penalty (ERP) \cite{chen2004marriage,chen2005robust}, Time Warp Edit distance (TWE) \cite{marteau2009time} and Move-Split-Merge (MSM) \cite{stefan2013move}.
A more complete description and comparison of these distance measures can be found in \cite{bagnall2017great,lines2015ee,wang2013experimental}. 
Note also that most of these distances have parameters of which tuning is critical to their functioning. 

\subsubsection{Nearest-neighbor approaches}
It is most common to 
classify time series data using Nearest Neighbour classification based on a relative distance (such as those returned from the measures above) \cite{Haghiri2017}. In fact, for more than a decade, the NN algorithm combined with the DTW measure was extremely difficult to beat \cite{wang2013experimental}. 

It is important to note here that when researchers mention the use of NN with a time series measure, the measure is not directly applied with a default parameterization, but rather its parameters are first learned on the data, usually by cross-validation. 

There are two main issues with NN approaches: (1) the tuning of the measures' parameters is usually quadratic with the size of the training data, and (2) the classification is at least linear with the size of the training data. Both of these issues are further compounded by the fact that most measures have a computational complexity that sits between linear and quadratic with the length of the series. 

To alleviate the second issue of scalable classification, targeted techniques have been developed. Data reduction techniques aim at simplifying the training database without penalizing the classification quality; either by directly removing objects from the original database \cite{pkekalska2006prototype,ueno2006anytime} or by summarizing the database and replacing sets of time series with representatives using average time series \cite{petitjean2012summarizing,petitjean2014dynamic,petitjean2016faster,marteau2016averaging}. 
Indexing is more difficult on time series than it is on traditional data, mostly because time series measures do not obey the triangular equality (at most obeying a relaxed $p$-triangular inequality \cite{lemire2009faster}), which makes exact pruning very inefficient (however, a general approach to improving the efficiency of NN searches in non-Euclidean space is available in \cite{Lifshits2010}).
To perform exact indexing, the main research effort has been put onto developing lower bounds (and mostly for DTW) \cite{keogh2006lbkeogh,lemire2009faster}. 
Recently, 
impelled by the motivating application 
\modif{of earth observation data analytics}, we have developed an algorithm for approximate and efficient NN search under DTW \cite{tan2017indexing}, an algorithm using the idea of a hierarchical $k$-means clustering.

As mentioned previously, Elastic Ensemble (EE) \cite{lines2015ee} is a recent state-of-the-art time series classifier. It is an ensemble of 11 NN classifiers, each learned with a different time series measure (with their parameters tuned accordingly). The EE algorithm has played a significant role in the design of Proximity Forest and will be discussed at greater depth in Sections~\ref{novelPF} and ~\ref{learnPF}.

\subsection{Approaches that learn features}
The following approaches construct an abstraction of the training dataset by learning features that represent the classes in the time series. 
\subsubsection{Shapelets} \label{shapelets}
The aim of shapelet algorithms is to find subseries (or consecutive subsets of time series) that can help discriminate between the different classes. 
To classify a time series, the learned shapelet is placed at the best position in the time series (usually under Euclidean distance), and the `matching' of the shapelet to the time series correspond to its distance at this best position. The original shapelet-classifier \cite{ye2011shapelets} inserted this algorithm at the node of a decision tree as a splitting criterion. 
This algorithm has a high training complexity ($O(n^4)$) due to the large number of candidate shapelets and the repeated scanning of the data. Subsequent research has focused on optimising the original algorithm to address both classification accuracy  and scalability, notably Fast Shapelets \cite{rakthanmanon2013fast}, Learning Time Series Shapelets \cite{grabocka2016fastshape} and Shapelet Transforms \cite{hills2014shapelets}.

Shapelet Transforms (ST) is a current state-of-the-art classifier that identifies the best $k$ shapelets in a single scan of the data (the number of shapelets can be reduced afterwards). The data is then transformed by defining an attribute to represent each shapelet with the value being the (usually Euclidean) distance between the shapelet and the best position in the time series. The transformed dataset can now be used with any classifier or ensemble of classifiers (such as in \cite{bagnall2015cote}). While ST is considered a state-of-the-art classifier, it has little potential to scale to large datasets given its training complexity of $O(n^2\cdot l^4)$. 

\subsubsection{Bag of Words approaches}
Bag of Words algorithms are similar to Shapelets in that they start by identifying exemplar subseries in the data to discriminate between classes. However rather than finding the similarity to the relative best positions in a time series, bag of words approaches differentiate classes by the relative frequency of the subseries. To calculate these frequencies, the algorithms discretise the values into a series of symbols, assigning letters to each subseries, and thus representing the original time series as `words'. Notable approaches are the Bag of Patterns \cite{Lin2012bop}; the Symbolic Aggregate Approximation-Vector Space Model (SAX-VSM) \cite{senin2013sax}; and the Bag of SFA Symbols (BOSS) \cite{schafer2015boss}, which is currently considered state-of-the-art.

The BOSS algorithm transforms the time series into a word using a Symbolic Fourier Approximation (SFA) \cite{Schafer2012SFA} thus making it robust to noise and delivering a high classification accuracy. It is however of limited use on large datasets as it has a high training complexity $O(n^2)$ \cite{bagnall2017great}. The authors identified this as a weakness and subsequently produced similar approaches with improved scalability, the Bag of SFA Symbols in Vector Space (BOSS-VS) \cite{schafer2015scalable}. The same authors recently proposed WEASEL \cite{Schafer2017WEASEL}, which improves on the computation time of BOSS and on the accuracy of BOSS-VS, but has a very high memory complexity (our experiments will show that it doesn't scale beyond 10,000 time series). In this way, WEASEL is more optimised for speed on small datasets than for scalability.


\subsection{Ensemble approaches}
Ensemble approaches are combinations of multiple classifiers. Each contributing algorithm can be weighted to maximize classification accuracy, while the time complexity is that of the slowest constituent. Some of these approaches have been discussed above as they are based around one main type of classifier, for example EE and ST.

The Collection of Transformation Ensembles (COTE) \cite{bagnall2015cote} is an ensemble comprising  35 classifiers across four time series domains: time, frequency, change and shapelet transformation.
For the time domain, COTE uses the 11 distance measures of EE, while in the other three domains, classifiers are recruited from outside time series classification -- $k$-nearest neighbours, naive Bayes, decision trees, random forest, rotation forest, support vector machines (two models) and a Bayesian network approach. On the benchmark UCR datasets, COTE has the highest average classification accuracy of all current approaches. However, its time complexity is bound by that of the Shapelet Transform, which is $O(n^2\cdot l^4)$ and the parameter searches for the elastic distance measures (EE), some of which are  $O(n^2\cdot l^3)$.

\modif{
\subsection{Decision Tree approaches}\label{sec:treeapproaches}

A number of decision tree approaches have been developed for time series classification.

Time Series Forest (TSF) \cite{deng2013TimeSeriesForest} first derives summary features for all time series by dividing them into intervals and summarising each interval by its mean, standard deviation and gradient. 
Then a Random Forest-like strategy is employed to select between a random subset of these features at each node in each of an ensemble of trees.  A novel selection criterion is used that considers both entropy gain and the margin by which a feature separates the classes.
This continues until the entropy gain ceases to improve, at which stage the node is defined as a leaf. TSF has been shown to be a reasonably accurate classifier: its accuracy ranks behind EE and DTW without being significantly worse \cite{bagnall2017great}. \modiff{However, its main virtue is computational efficiency. TSF learns in $O(n \log(n)\cdot l \cdot k)$ for a forest of $k$ trees built from $n$ series of length $l$, which is a much lower complexity than the current state-of-the-art.}

Generalized Random Shapelet Forest (gRSF) \cite{karlsson2016ShapeletForest} extracts a shapelet from
a randomly chosen time series and finds the distance between this time series and each other time series. The data is then split according to whether it is above or below a threshold distance to the representative shapelet. This is applied recursively until the node is either pure on the number of instances remaining at a node is less than 3. As mentioned in section \ref{shapelets}, the main pitfall of shapelet-based methods is the high computational cost of finding candidate shapelets and comparing shapelets to other time series. The gRSF minimises this issue by randomising many of the model choices---a candidate shapelet is generated from a randomly chosen time series by choosing a random starting point and random length, this is repeated $r$ times and the best candidate is chosen for a given split. The resulting algorithm has accuracy competitive with Learning Time Series Shapelets  and better than DTW.

\modif{A number of approaches have been developed that form decision trees where splits are based on similarity to chosen exemplars \cite{Balakrishnan2006,Douzal2012treetimeseries,Yamada2003decision}. One strategy is to select a single exemplar and then choose a cut point on a similarity measure with respect to that exemplar. Series with similarity scores lower than the cut point follow one branch and the remaining examples follow the other.  The other strategy is to select multiple exemplars, one associated with each branch. Series follow the branch with whose exemplar they are most similar.
These approaches are hampered by the high computational complexity of their search for exemplars at each node.  
Similarity Forests \cite{Sathe2017SimForest} and Comparison-based Random Forests \cite{Haghiri2018} generalise this idea to attribute-value data with random selection of exemplars and developed forests of such trees. Similarity Forests add a cutoff value on the difference in the distance between the two exemplars, and optimizes that cutoff value based on weighted Gini.}
The idea of using similarity as the splitting criterion in tree structures has also been successfully used for \emph{indexing} of regular tabular data \cite{flann,annoy,Sathe2017SimForest} and of time series with DTW \cite{tan2017indexing}.

}

\section{Proximity Forest}
\label{sec:Proximity Forest}
In this section, we present our novel algorithm for time series classification: Proximity Forest. We start by highlighting why there is a need for a new time series classifier. We then present our model and the two key algorithms (1) how to learn a Proximity Forest and (2) classifying with a Proximity Forest. We conclude this section with some comments about its complexity. 

\subsection{Why do we need a novel time series classifier?}
\label{novelPF}
The previous section highlighted that the last decade has seen numerous classifiers and distance measures specifically designed for time series classification. Based on this, one could wonder why there is a need for a novel algorithm; the answer is simple: most state-of-the-art algorithms do not scale to large time series datasets. We have seen that some do not scale in the learning phase (ST, EE, COTE). Others require a scan of the training database to perform each classification (EE, COTE). Those that do scale to medium-size datasets, such as BOSS-VS, compromise accuracy in order to do so (as we will show for both our case study and for UCR datasets). Throughout the development and advancement of much of the current state-of-the-art, scalability has usually been secondary to accuracy. This is because time is not a significant issue when considering data with only few time series. However,
\modif{a growing number of modern applications
consist of hundreds of thousands to millions of time series.} 
These applications require a classifier that is both accurate \emph{and} scalable in both learning and classification.

BOSS-VS is a classifier that appears to have developed with a focus on scalability. However, as we will see in Section~\ref{sec:Experiments}, its accuracy ranks some 30 percentage points lower in our case study, and therefore is not competitive with the accuracy of the state-of-the-art.

While COTE is currently \emph{the} state-of-the-art \modif{in terms of accuracy}, its learning phase is bound by the runtime complexity of both ST and EE.
On our 1M dataset --~and as depicted in Figure~\ref{fig:intro-train-time}~-- the sole learning phase of COTE associated to training EE would require \textbf{73 thousand}  days, or 200 years. This is even more startling knowing that the series in this dataset are very short with only 46 timestamps. 

\modif{The large runtime complexity of COTE is largely due to the fact that EE does not abstract much information during the learning phase, and therefore has a significantly greater number of processes to complete during testing. A corollary of this is that a distance-based classifier that learns faster than EE for the same level of accuracy would also present an improvement to COTE. It is for this reason that our design of algorithm incorporates many elements of EE---11 distance measures and similar parametrisation---and why our experiments provide a direct comparison of Proximity Forest against EE.}

We therefore argue that the need for a scalable and accurate classifier has not yet been met.
We incorporate three critical elements into the design of our novel approach: 
\begin{enumerate}
\item We make the most of over 30 years of research into designing consistent measures for time series.
\item We specifically design our ensemble to have a high variability between the different individual classifiers. This results in an improved overall classification accuracy over a single classifier, based on the principle of ensemble methods. In general, averaging the predictions of multiple models each having high variance and low bias results in an ensemble classifier with a lower total error than any single classifier. This is analogous to how a Random Forest model, another ensemble of decision trees, will only learn from a fraction of the available features for each individual node in order to introduce variability between the trees \cite{Breiman2001RF,Ho1995Rdf}. This observation is important, because we did not design the learning of an individual tree to maximize its accuracy; if we had wanted to design a single tree model, we would have made different design choices. We designed the learning of individual trees so that the overall classification performance is maximised. 
\item We design Proximity Forest to be extremely scalable with an \modif{average-case learning complexity of $O(n \log\left( n\right) \cdot l^2)$ and a classification complexity of $O(\log \left(n\right)\cdot l^2)$ per tree for $n$ training time series of length $l$. This contrasts with the state of the art learning in $O(n^2\cdot l^3)$ (Elastic Ensemble) or $O(n^2\cdot l^4)$ (Shapelet Transform, COTE). Again here, we might have made different choices if scalability wasn't a design objective.}
\end{enumerate}
\modif{To achieve scalability we employ tree-based classifiers. These are scalable due to their use of a divide-and-conquer strategy.  At each level the data are divided into multiple subsets, as result of which the trees are on average of depth $O(\log n)$, hence increasing sublinearly in depth relative to training set size.}

\modif{Our use of decision trees for time series classification is not novel in itself. Trees are attractive due to their divide and conquer methodology and resulting potential for efficient learning and classification. Previous implementations, however, have lacked competitiveness in accuracy \cite{deng2013TimeSeriesForest,Douzal2012treetimeseries,Yamada2003decision} or time \cite{Balakrishnan2006,Douzal2012treetimeseries,Yamada2003decision}. 
}

\modif{To achieve scalability we merge the strategy of learning decision trees where splits are based on similarity to chosen time series exemplars \cite{Balakrishnan2006,Douzal2012treetimeseries,Yamada2003decision} with the strategy of forming forests of such trees in which the exemplars are chosen at random \cite{Sathe2017SimForest}. To this amalgam we add the critical ingredient of stochastic selection between a large range of similarity measures, which both reduces bias and provides a beneficial increase in variance between ensemble members.}



\subsection{How to learn a Proximity Forest?}
\label{learnPF}
We seek to learn a Proximity Forest from a training set comprising $n$ labeled time series, each of which is  of length $l$, where the labels are integers from $1$ to $c$.

A Proximity Forest is an ensemble of $k$ Proximity Trees.
A Proximity Tree is similar to a regular decision tree, but differs in the tests applied at internal nodes. Whereas a regular decision tree applies a test based on the value of an attribute (e.g.\ if height > 160 cm, follow the left branch, otherwise follow the right branch), each \emph{branch} of an internal node of a Proximity Tree has an associated exemplar 
and an object follows the branch corresponding to the exemplar to which it is \emph{closest} according to a parameterized similarity measure. We will see later how the exemplars and measures are chosen. 
A~tree is either a leaf or an internal node.

An internal node has two fields, \texttt{measure}, a function \modif{$\mathit{object}\times \mathit{object} \rightarrow \mathbb{R}$}, and \texttt{branches}, a vector of branches. Each branch has two fields, a time series (\texttt{exemplar}) and a tree to which an object is passed if it is nearer to the branch's exemplar than any other (\texttt{subtree}).

\modiff{
If all data reaching a node has the same class, i.e. is pure, the $create\_leaf$ function creates a new leaf node and assigns this class label to its field \texttt{class}. This label is then assigned to any query time series reaching this leaf during the testing phase.
}

\paragraph{How do we choose the splitting criteria?} A Proximity Tree creates, at each node, one branch for each class that exists in the data it receives from its parent. These exemplars are chosen uniformly at random among each class. The parameterized similarity measures are also chosen uniformly at random among a pool that will be described below after we have given the main overview of the algorithm. We will detail, after the main algorithm, how  it is possible to \emph{learn} with randomized trees. 

Algorithm~\ref{build} presents the algorithm for learning a single tree. Each node is constructed recursively from the root node down to the leaves. If the data at the node is pure --~ie. all data belongs to the same class~-- then the node becomes a leaf and the recursion finishes. 

\modif{At each node, a pool of $r$ candidate splits are evaluated (Algorithm~\ref{newsplitter}). For each candidate, a parameterised measure is chosen uniformly at random among a pool 
of such measures. We then select an exemplar for each class represented at the node and pass the data down the branches by finding the closest exemplar (one per class) for each time series in the data using the split's distance measure. 

Once each candidate split has been created, we then select the candidate that maximizes the difference between the Gini impurity of the parent node and the weighted sum of Gini impurity of the child nodes. We then call the construction of the tree recursively on each branch for the successful candidate; this constructs all subtrees. When this is done, the tree is constructed.

Increasing the number of candidate splits per node will lead to an improvement of the quality of each split. However, it will also lead to an increase of the training time. The choice for the value of $r$ will be discussed later in Section~\ref{subsubsec:candidates_eval}.
}

\modiff{
  \begin{algorithm}
    \caption{$\mathrm{build\_tree}(D,\Delta,R)$}
    \label{build}
    \DontPrintSemicolon
    \KwIn{$D$: a time series dataset}
    \KwIn{$\Delta$: a set of parameterized distance measures}
    \KwIn{$R$: number of candidate splits to consider at each node}
    \KwOut{$T$: a Proximity Tree}
\vspace*{5pt}
    \uIf{$\mathrm{is\_pure}(D)$}{\Return $\mathrm{create\_leaf}(D)$\;
	}
\vspace*{5pt}
	\tcp{create tree, to be returned, represented as its root node}
    $T \leftarrow \mathrm{create\_node}()$\;
\vspace*{5pt}
    \tcp{Creating $R$ candidate splitters}
    $\mathcal{R} \leftarrow \emptyset$\;
    \For{$i=1$ \KwTo $R$}{
      $r\leftarrow gen\_candidate\_splitter\left(D,\Delta\right)$ \tcp*{generate random splitter}
      Add splitter $r$ to $\mathcal{R}$\;
    }
\vspace*{5pt}
    \tcp{select best splitter; it splits using measure $\delta^\star$ and exemplars $E^\star$}
    $(\delta^\star,E^\star)\leftarrow \argmax\limits_{r\in \mathcal{R}}  \mathrm{Gini}\left(r\right)$\;
    
\vspace*{5pt}
    $T_\delta \leftarrow \delta^\star$ \tcp*{retain measure for root node of $T$}
    $T_B\leftarrow \emptyset$\tcp*{$T_B$ will store the branches under root node of $T$}
    \ForEach{\emph{exemplar} $e \in E^\star$}{
      \tcp{$D^\star_e$ is the subset of $D$ that are the closest to $e$ based on $\delta^\star$}
      $D^\star_e \leftarrow \left\{d\in D \;\raisebox{-4pt}{\resizebox{2.5pt}{20pt}{$\mid$}}\; \argmin\limits_{e'\in E^*}\delta^\star(d, e') = e\right\}$\;
      $t\leftarrow \mathrm{build\_tree}(D^\star_e,\Delta,R)$ \tcp*{build subtree for that branch}
      Add branch $\left(e, t\right)$ to $T_B$\tcp*{a branch is a pair $(\text{exemplar,sub-tree})$}
    }
\vspace*{5pt}
    \Return $T$
\end{algorithm}
}
\modiff{
  \begin{algorithm}
    \caption{$\mathrm{gen\_candidate\_splitter}(D,\Delta)$}
    \label{newsplitter}
    \DontPrintSemicolon
    \KwIn{$D$: a time series dataset.}
    \KwIn{$\Delta$: a set of parameterized distance measures to sample from}
    \KwOut{$(\delta, E)$: a parameterized distance measure and a set of exemplars}
\vspace*{5pt}
    $\delta \xleftarrow[]{\sim}\Delta$ \tcp*{sample a parameterized measure $\delta$ uniformly at random from $\Delta$}
    \vspace*{5pt}
    \tcp{Select one exemplar per class to constitute the set $E$}
    $E \leftarrow$ $\emptyset$\;
    \ForEach{\emph{class} $c$ \emph{present in} $D$}{
      $D_c \leftarrow \left\{d\in D \mid class(d) = c\right\}$ \tcp*{$D_c$ is the data for class $c$}
      $e \xleftarrow[]{\sim}D_c$ \tcp*{sample an exemplar $e$ uniformly at random from $D_c$}
      Add $e$ to $E$
    }
\vspace*{5pt}
    \Return $(\delta, E)$
  \end{algorithm}
}

\paragraph{The case of $R=1$: is selecting at random still `learning'?} One might wonder what the tree is actually learning when one only considers a single candidate ($R=1$). In that case, no selection of `the best possible split' is performed. It is interesting to note that choosing splitting criteria independently of the output value has been studied before, a key example being Extremely Randomized Trees \cite{geurts2006extremely}. In that work, they showed that splitting completely at random still ensures consistency (tending to Bayes Optimal error as the data tends to infinity). 
The main reason is that the exemplars are not random points in the input space. They are sampled from the data distribution of each class. In~consequence, the trees \emph{are} still learning an abstraction of the data, using the trees as a density estimator \cite{ting2016overcoming}.

\begin{figure}
\centering
\includegraphics[width=.95\linewidth]{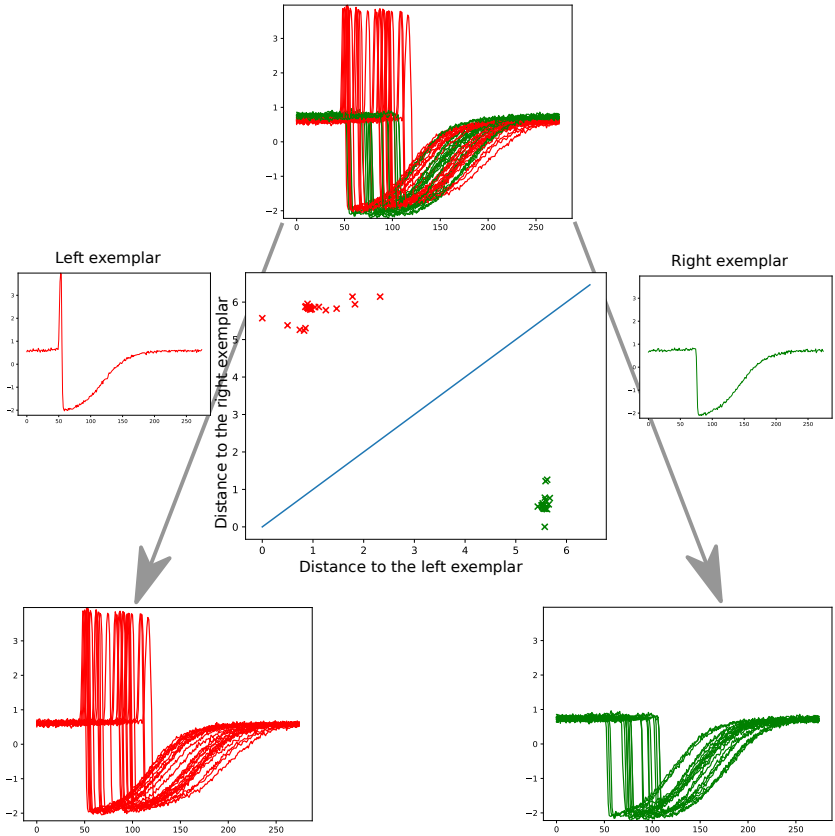}
\caption{Visual depiction of the root node for the `Trace' dataset (simplified to 2 classes). The top chart represents the data at the root node (one colour per class) while the data at the bottom left and right represent the data once split by the tree. The two time series in the middle left and right are the exemplars on which the tree is splitting. The scatter plot at the center represents the distance of each time series at the root node with respect to the left and right exemplars (resp. x- and y-axes). }
\label{fig:splitting}
\end{figure}

We depict in Figure~\ref{fig:splitting} a graphical representation of a simple split obtained on the \texttt{Trace} dataset. It is interesting to see that \modiff{
in Euclidean space, the splitting criterion is actually forming a hyperplane that is equidistant to the exemplars. Note that this intuition is more complex for time series measures, because most of them do not have properties of a metric \cite{lemire2009faster}. The scatter plot depicts each time series as a dot in this space, with the x-axis representing distance to the first exemplar and the y-axis distance to the second.}

\paragraph{How to choose the parametrised measure on which to split?}
The parametrised distance measure gives a measure of the similarity between the exemplar time series. \modif{For each candidate split at each node}, the algorithm chooses a distance measure at random from the following 11 distance measures used by the Elastic Ensemble (EE) learner that we described above: Euclidean Distance (ED); Dynamic Time Warping using the full window (DTW); Dynamic Time Warping with a restricted warping window (DTW-R); Weighted Dynamic Time Warping (WDTW); Derivative Dynamic Time Warping using the full window (DDTW); Derivative Dynamic Time Warping with a restricted warping window (DDTW-R); Weighted Derivative Dynamic Time Warping (WDDTW); Longest Common Subsequence (LCSS); Edit Distance with Real Penalty (ERP); Time Warp Edit Distance (TWE); and, Move-Split-Merge (MSM). Randomising the choice of distance measure is a deliberate decision to introduce variability between each tree, for the reasons stated earlier.

Once a distance measure is chosen at random, it is then parametrised. 
\modif{The parametrisations are addressed in turn. They are deliberately chosen to mimic as closely as possible the EE algorithm. Even though better values might be chosen here, we mimic EE's parameterization to allow direct comparison. Euclidean distance, full DTW, and full DDTW distances have no parameters to select. DTW-R and DDTW-R only require a warping window parameter that is chosen uniformly at random in $[\!\![0,\lfloor\frac{l+1}{4}\rfloor]\!\!]$ (thus allowing a warping of elements at most $\frac{l}{2}$ apart). WDTW and WDDTW requires also one parameter to select that it is used into the weighted value $g$ to control the level of penalization between two different time stamps -- we use $g\sim U(0,1)$. The parametrisation of ERP is a distance threshold that controls for how close elements have to be to be considered similar; we sample it uniformly at random in $[\frac{\sigma}{5},\sigma]$, with $\sigma$ being the standard deviation of the data. LCSS has as first parameter the same distance threshold value (which is sampled in the same way), and has a second parameter --~the warping window size~-- which is chosen in the same way as for DTW-R. TWE has two parameters $\gamma$ and $\lambda$ which respectively control for the stiffness and penalty value in the alignment. Following \cite{marteau2009time}, $\lambda$ is sampled at random from $\cup_{i=0}^{9} \frac{i}{9}$ and $\gamma$ following at random from the exponentially growing sequence $\{10^{-5},10^{-4},5\dot 10^{-4},10^{-3},5\dot 10^{-3},\cdots,1\}$, resulting in 100 possible parameterizations. The final measure, MSM, has a single parameter which is sampled from an exponential sequence similar to the one for $\gamma$ in TWE with 100 values ranging from $10^-2$ to $10^2$, as recommended in \cite{stefan2013move}.

Choosing the parameter at random has a twofold effect: 1) it skips the cross-validation step which has a quadratic complexity; and 2) it introduces variability between trees, which provides superior learning through lower-biased trees and ensembling.
}
In the following experiments we will show that Proximity Forest is not only orders of magnitude faster than EE, but that its accuracy also ranks higher than EE. 

\subsection{Classifying with a Proximity Forest}
The process of classification for a single Proximity Tree is detailed in Algorithm~\ref{alg:classification}: a query time series begins at the root node and the distance from the query to each of the exemplar time series is calculated, by using the node's distance measure and exemplars selected when constructing the tree. The query time series is then passed down the branch of the exemplar to which it is nearest. The query time series then traverses down the tree by repeating this process until it reaches a leaf, where it is assigned the class represented by that leaf. This process is repeated for each tree constructed as part of the forest. A~Proximity Forest then uses majority voting between its constituent Proximity Trees.

\begin{algorithm}
  \caption{classification$\left(Q, T\right)$}
  \label{alg:classification}
  \DontPrintSemicolon
  \KwIn{$Q$: Query Time Series}
  \KwIn{$T$: Proximity Tree}
\vspace*{5pt}
  \uIf{$\mathrm{is\_leaf}(T)$}{\Return majority class of $T$}
\vspace*{5pt}
  \tcp{find the branch with exemplar closest to $Q$ using measure $T_\delta$}
  $(e, T^\star) \leftarrow \argmin\limits_{(e',T')\in T_B} T_\delta(Q,e')$\;
  
\vspace*{5pt}
  \Return classification$\left(Q, T^\star\right)$ \tcp*{recursive call on subtree $T^\star$}
\end{algorithm}

\subsection{Comparative complexity analysis}
\label{subsec:computation_analysis}

\modif{
 During the training phase, at each node, let us assume that $n'$ data points are present at the node. We first scan it once taking $O(n')$ time to split the data into $c$ groups, one for each class $c$ present at the node.
}

\modif{We then generate $r$ candidate splits, \ie{}$r$ sets of exemplar time series. For each such candidate set, we sample $c'\leqslant c$ exemplar time series, \ie{}one time series for class represented among the $n'$ time series available at the node -- this is done in $O(1)$ given that the data is already organised by class. For the candidate split to be operational, we also require a parameterized measure to use to compare against these $c'$ exemplars. Most of the parametrized measures can be chosen in $O(1)$, except for LCSS and ERP which calculate the standard deviation in $O(n'\cdot l)$ for data at the node while selecting the parameter.\footnote{\modif{Note that these parametrisations can be performed in constant time also if the data are $z$-normalized, which is the case for all UCR datasets.}}}

\modif{We now have $r$ candidate splits that are ready to be evaluated. We now wash the $n'$ time series down the branches for all candidate splits. This is done by comparing each time series to the $c'$ exemplars, each comparison taking from $O(c\cdot l)$ to $O(c\cdot l^2)$. Overall, this takes $O(n'\cdot c'\cdot l^2)$. If $r=1$, the training process at this node is finished and we call the training function recursively for each of the $c'$ children nodes. If $r>1$, we calculate the Gini coefficient for each of the $r$ candidate splits in $O(c^2)$, keep the best one, and delete other candidate splits.}

\modif{As the total number of examples that reach any of the nodes at a single given level cannot be greater than the total number of examples, $n$, the total computation per level of the tree is thus $O(n\cdot r\cdot c\cdot l^2)$. In the worst case, the majority of the training data at each level will pass down a single branch and the depth of the tree will be $O(n)$, resulting in a worst training time complexity of $O(n^2\cdot r\cdot c\cdot l^2)$. However, as the exemplars are following the class distribution, unless the data are in some way degenerate (for example if one class comprises only outliers), the average tree depth can be expected to be $O(\log n)$. In practice it will often be much smaller, because, unlike typical divide and conquer approaches, the tree terminates as soon as a node is pure rather than having to separate each individual object. Thus, for non-degenerate data we can expect average case training time complexity of $O\left(n\log\left( n\right) \cdot r\cdot c\cdot l^2\right)$ for a single tree and thus $O\left(k\cdot n\log\left(n\right) \cdot r\cdot c\cdot l^2\right)$ for a full Proximity Forest comprising $k$ Proximity Trees.}

The experiments presented in the next section will include runtimes and comparison to current state-of-the-art algorithms. These confirm that this expected average case quasi-linear complexity with respect to data quantity is borne out in practice.

During classification, a time series of length $l$ will pass through an average of $\log n$ nodes on each of the $k$ trees. At each node, the distance to at most $c$ exemplars must be computed. For each of these distance computations, the complexity will again depend upon the chosen distance measure;  the fastest being $O(l)$ and the slowest $O(l^2)$. Thus, the resulting average case complexity is $O(k\cdot \log n \cdot c \cdot l^2)$.

\section{Experiments}
\label{sec:Experiments}
This section describes the experiments that evaluate our Proximity Forest.
\modif{We start with} the Satellite Image Time Series (SITS) dataset, a (very) large time series dataset describing the evolution of the Earth as pictured every five days by a high-resolution satellite. This dataset \modif{is an example of the large time series datasets that motivate} the need for a new time series classification algorithm, as no current state-of-the-art approach scales to this magnitude. Conversely, there are classifiers designed for scalability, namely BOSS-VS, that compromise classification accuracy to do so. The first experiments presented in this section use the SITS dataset to demonstrate the ability of Proximity Forest to be both scalable \emph{and} accurate. The second section assesses the Proximity Forest on the datasets of the UCR time series classification repository \cite{UCR}, the benchmark in the field. It demonstrates that the classification accuracy of Proximity Forest is competitive with the current state-of-the-art. The final section discusses other considerations surrounding Proximity Forest, such as the effect of varying the number of trees, and the standard deviation of the results.

It should be mentioned that throughout the following experiments we have emphasized a comparison with EE. This is because it is viewed as the closest relative to Proximity Forest amongst the current state-of-the-art, given that neither method includes data transforms or shapelets. It is also the constituent of COTE that bounds its learning time and therefore any improvement over the runtime of EE, for the same classification accuracy, would also equate to an improvement on COTE, the current leader in the field.

To facilitate others to build on our work, as well as to ensure reproducibility, we have made our code and the full raw results available at \ourRepo{}. 

\subsection{Case study: Satellite Image Time Series Dataset}
\label{subsec:sits_expe}
The SITS dataset contains approximately 1 million time series with a train-test split of approximately 90\%-10\%.\footnote{The split ensures that no 2 times series come from the same plot of land.} Each time series has a length of 46 and is labeled as one of 24 possible land-use classes (e.g.\ `wheat', `corn', `plantation', `urban'). \modif{Here the labeled data has been extracted from three sources: 1) ground field campaigns for most of the vegetation classes, 2) farmer's declaration to complete the data for some crop classes, and 3) existing map for the urban areas.}

The experiments presented in this section were performed on this dataset, comparing the performance of Proximity Forest against three competitors: BOSS-VS (designed for scalability), WEASEL (designed for speed and quality), and EE (designed for quality). We use 5 runs for each experiment of Proximity Forest and 1 run of each of the competitors -- as their results are deterministic.
Throughout this experiment, we use 100 trees; we will see in Section~\ref{subsec:pf_variations} that this gives a good tradeoff between accuracy and computational time/memory. \modif{Although we are mainly assessing the scalability, we will also have a quick look to the accuracy}.

\subsubsection{Training scalability}
\begin{figure}
\centering
\subfigure[]{\includegraphics[width=.46\linewidth]{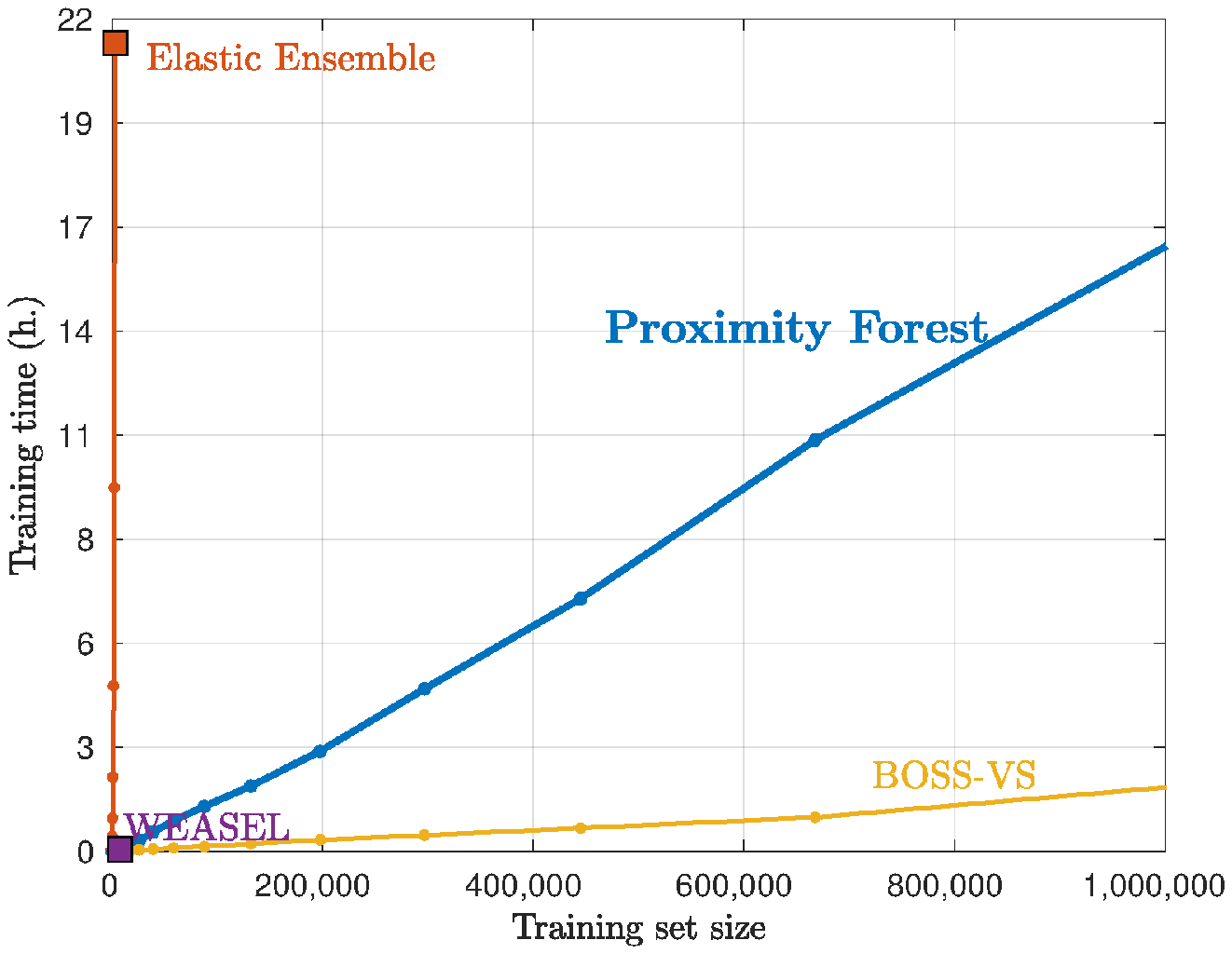}}
\subfigure[]{\includegraphics[width=.46\linewidth]{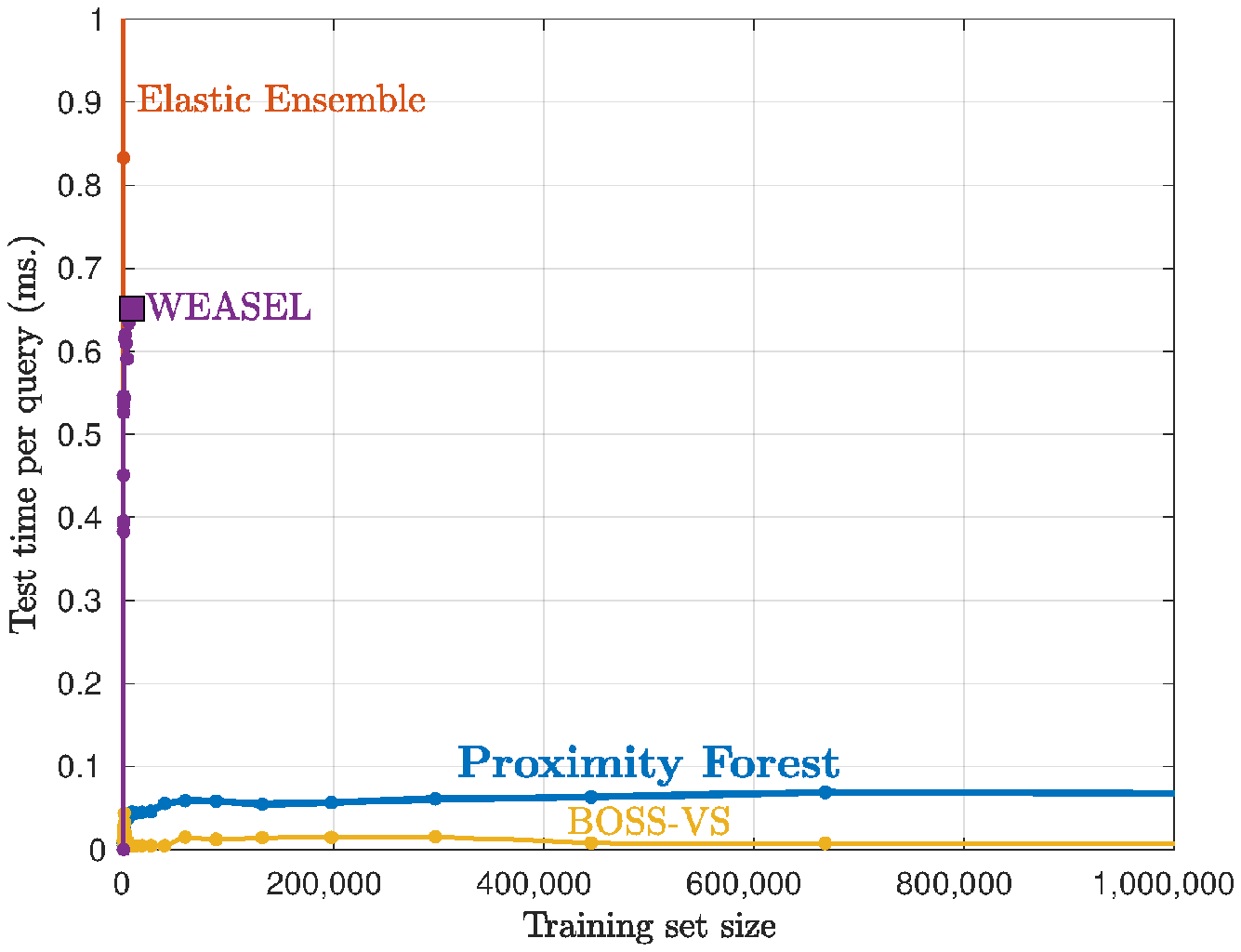}}
\caption{Training time (a) and testing time per query (b) as a function of training set size for Proximity Forest, EE, WEASEL and BOSS-VS.}
\label{fig:SITS-train-test}
\end{figure}

To assess scalability, we train and test each algorithm on subsample data with increasing training set size, allowing training time, testing time and accuracy to be measured as a function of training size. Figure~\ref{fig:SITS-train-test}(a) shows training time against training size for each of the 4 algorithms. 
\paragraph{Versus EE.}First, it is evident that Proximity Forest presents a notable saving in training time over EE, confirming that it trains in linear time rather than the quadratic time for EE. Even for a small training set of about 2,000 time series, learning an EE model took about 10 hours, compared to Proximity Forest's 79 seconds. Fitting a quadratic curve through both EE and Proximity Forest is quite informative: EE returns a quadratic component of 6.3 while Proximity Forest only \num{-8.10e-6}, clearly highlighting both the quadratic complexity of EE, and also that Proximity Forest is in practice very close to its theoretical average complexity and scales quasi-linearly with $n$.
\paragraph{Versus WEASEL.}
WEASEL is very fast but its memory footprint did not allow it to scale beyond 8,000 time series even when given 64 GB of RAM. This clearly highlights the difference with BOSS-VS: we can see that WEASEL was not developed for scalability, but rather for speed on small datasets.
\paragraph{Versus BOSS-VS}Proximity Forest trains slower than BOSS-VS for a given training size, however this is counteracted by \modif{the low accuracy of BOSS-VS} discussed below. 

\subsubsection{Testing scalability}
\modif{Figure~\ref{fig:SITS-train-test}(b) shows testing time against training size for each of the 4 algorithms. }
The story here is very similar to that of training: it confirms the way Proximity Forest scales logarithmically with training set size, while EE must scan the full database many times. Here again, WEASEL becomes infeasible to apply with relatively small quantities of training data. Proximity Forest and BOSS-VS require respectively 0.0679 ms and 0.0077 ms to classify a time series with a model trained on 1M time series. 

\subsubsection{Is Proximity Forest accurate and scalable?}
\begin{figure}
\centering
\includegraphics[width=.65\linewidth]{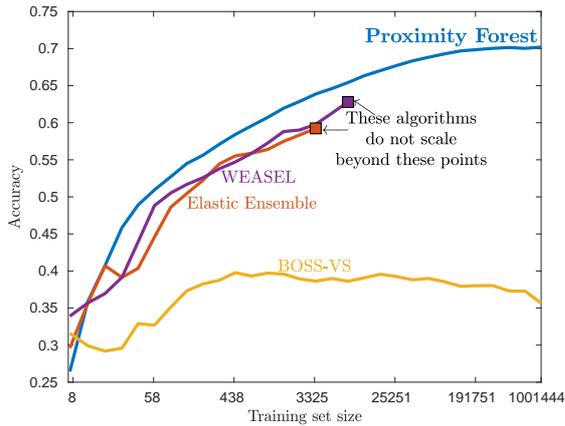}
\caption{Accuracy as a function of training set size for Proximity Forest, EE, WEASEL and BOSS-VS.}
\label{fig:SITS-accuracy}
\end{figure}

We have now seen that Proximity Forest is highly scalable and only beaten by BOSS-VS in terms of training time. We will now study how its accuracy scales with training set size. 
The main results are presented in Figure~\ref{fig:SITS-accuracy} which plots the accuracy as a function of training set size for Proximity Forest, EE, WEASEL and BOSS-VS. 

The first element to observe is that Proximity Forest obtains \modif{greater accuracy than the competitors} for large training sets. WEASEL and EE become infeasible to apply at relatively small data quantities and BOSS-VS --~which is faster than Proximity Forest~-- does not learn effective classifiers on this dataset. With 
63.8\% accuracy at 3,400 training set size, this is 
26.3 percentage points more accurate than BOSS-VS, and  
4.6 and 4.7 percentage points more accurate than WEASEL and EE, respectively. \modif{Such differences are substantial in a problem comprising 24 classes.} 

\modif{Moreover, Proximity Forest is more accurate than the other algorithms from 500 training instances upwards}. This is not surprising, as trees usually have a better control over variance than NN algorithms, because of their higher bias and abstraction capabilities. Proximity Forest thus appears to be \modif{both accurate and highly scalable. We will show in the next subsection that this result holds also on the benchmark UCR archive.}

\subsection{Experiments on the UCR Archive}
In this section, we study the behavior of Proximity Forest on the 85 datasets of the traditional UCR archive \cite{UCR}. 
It is useful to remember here that our aim is not to show that Proximity Forest is more accurate than the state-of-the-art, but only that it is competitive while being highly scalable. 
We compare the mean error-rate of Proximity Forest to the error-rates on the standard train/test split for the state-of-the-art, as tested in \cite{bagnall2017great}. \modif{We average Proximity Forest results over 10 runs for each experiment.}
We compare Proximity Forest to five classifiers currently representing the state-of-the-art -- DTW-R, COTE, EE, ST and BOSS\footnote{It should be highlighted that the results presented here are for the original BOSS algorithm, and not the BOSS-VS discussed above in the SITS experiments. BOSS-VS is a scalable variation of BOSS, where concessions are made to accuracy in favor of training time. The original BOSS is therefore more competitive in this section.}.
\modif{The Proximity Forest results are obtained for 100 trees with selection between 5 candidates per node. A detailed discussion about the Proximity Forest parameters will be performed in Section \ref{subsec:pf_variations}.}

\begin{figure}
\centering
\subfigure[]{\includegraphics[width=.48\linewidth]{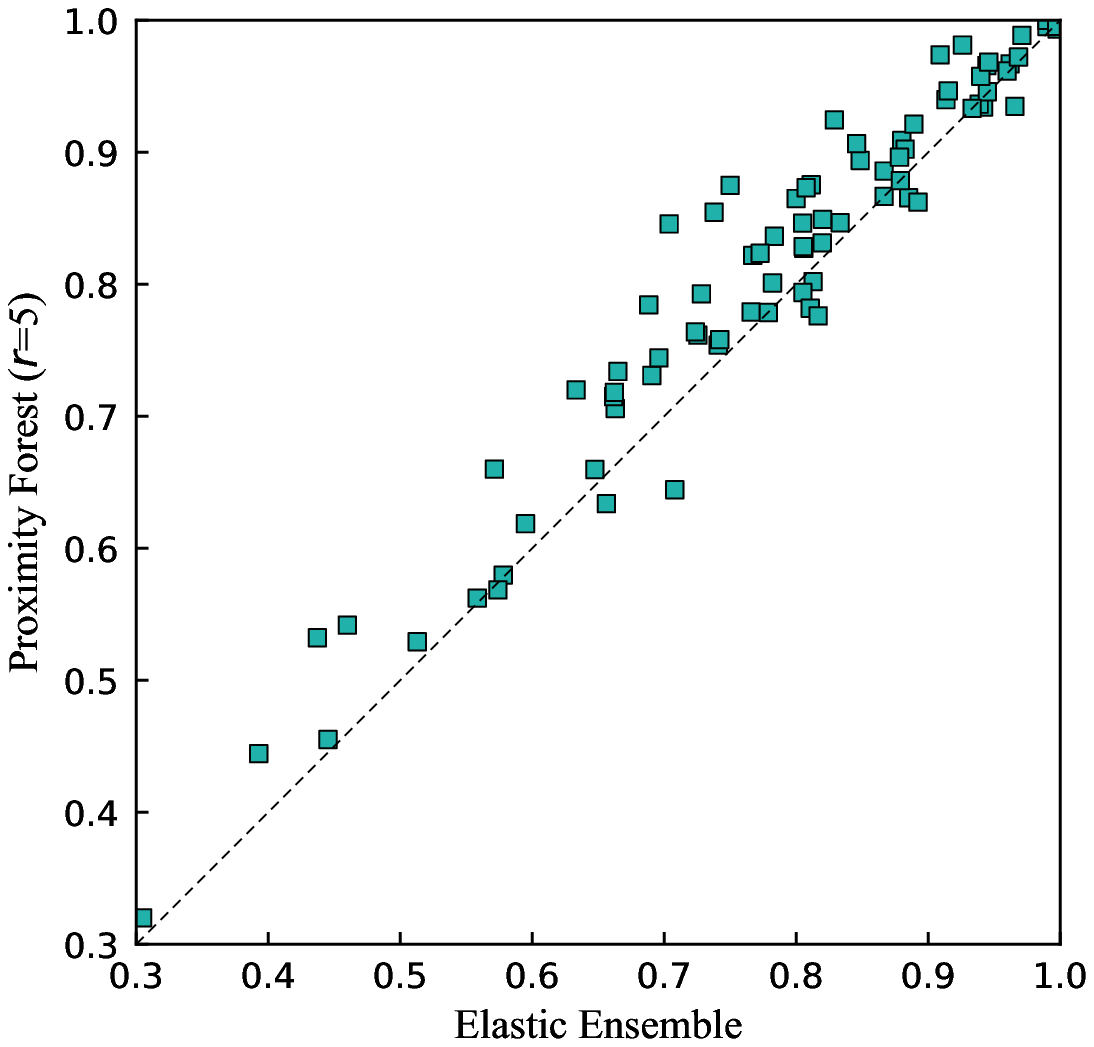}}
\subfigure[]{\includegraphics[width=.48\linewidth]{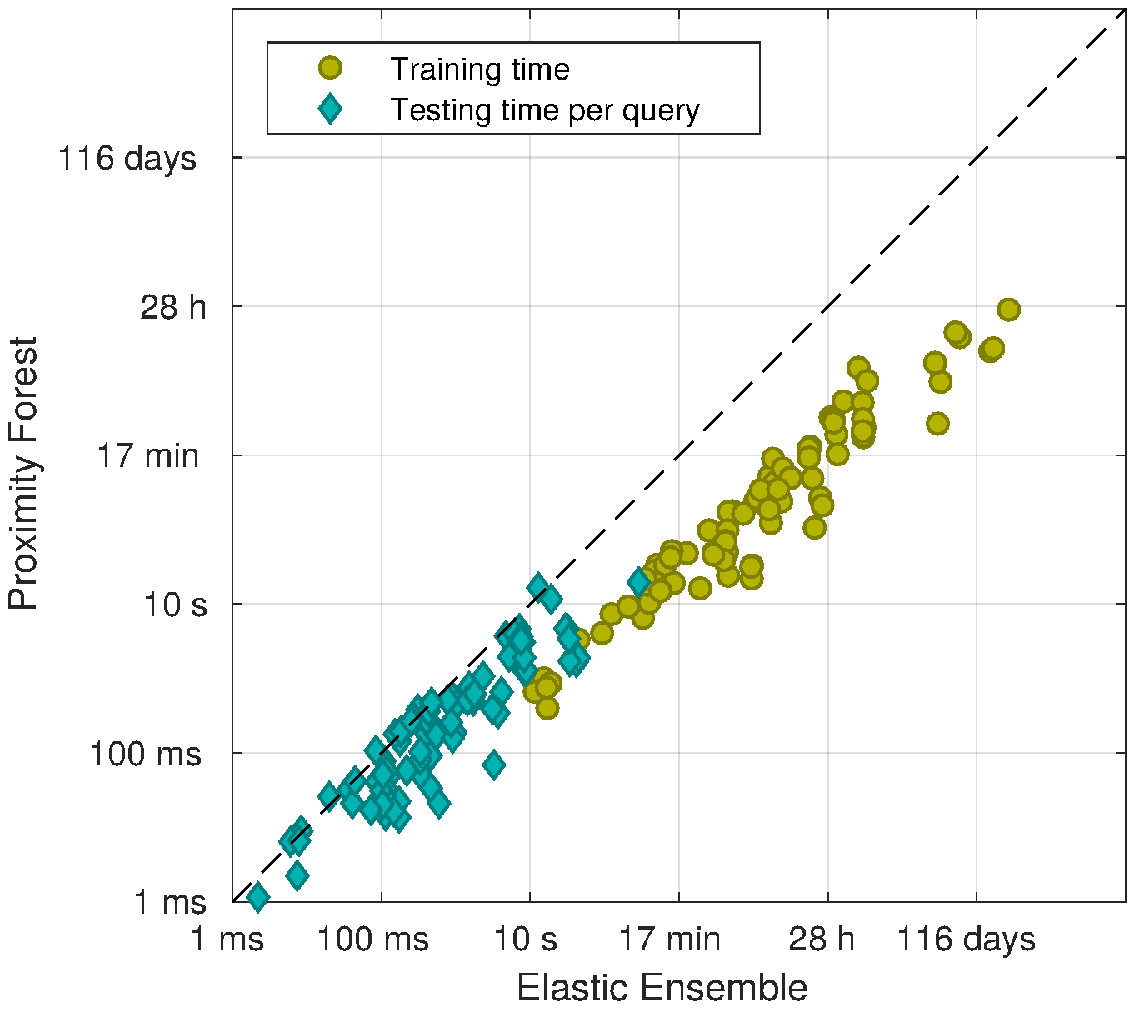}}
\caption{Comparison of Proximity Forest and Elastic Ensemble classifiers on UCR datasets in terms of a) accuracy and b) training and testing times in log scale.}
\label{fig:acc_EE_PF_UCR}
\end{figure}

We first show the comparison with Proximity Forest's closest relative, EE. Figure \ref{fig:acc_EE_PF_UCR}%
provides scatter plots of the relative accuracy, total training time and total testing time of each of these classifiers. Each point represents a different UCR dataset.
\modif{
Figure~\ref{fig:acc_EE_PF_UCR}(a) shows that Proximity Forest is more accurate on 
\modif{60} datasets and less accurate on  
only \modif{11} datasets, with \modif{14} ties. Moreover, for many datasets Proximity Forest is substantially more accurate than EE.}

Figure~\ref{fig:acc_EE_PF_UCR}(b) demonstrates that Proximity Forest has several orders of magnitude advantage in training time. When considering testing time, Proximity Forest has greater test time per query than EE for  
\modif{12} datasets, the majority of which are small datasets (\textit{i.e.} less than \modif{50} training instances). The largest such difference is observed for the Phonemes dataset for which Proximity Forest takes about 
\modif{17 seconds} per query compared to 13 seconds per query for EE. In contrast, the test time for Proximity Forest is much smaller than EE for the biggest datasets (\textit{i.e.} more than 800 training instances). For example, the biggest test time difference is for the HandOutlines dataset for which Proximity Forest takes about 
\modif{19 seconds} per query compared to 286 seconds per query for EE.

\begin{figure}
\centering
\includegraphics[width=.95\linewidth]{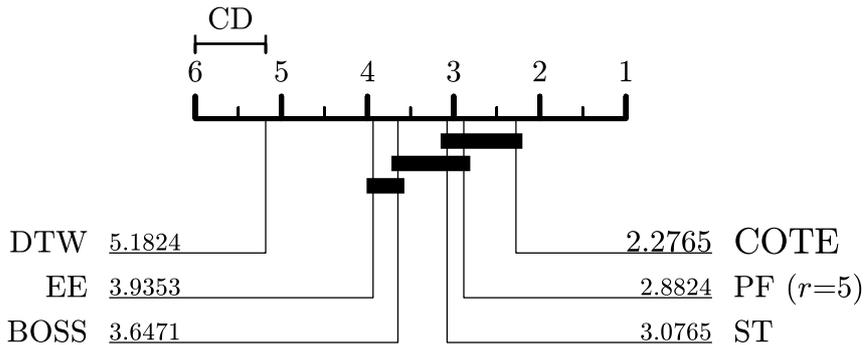}
\caption{Critical difference diagram for five state-of-the-art classifiers and Proximity Forest (PF) with 5 candidates.}
\label{fig:CriticalDiffDiagram}
\end{figure}

The commonly accepted method to compare multiple classifiers over multiple datasets is by average ranks. For each dataset, we rank the classifiers and then calculate the average of each classifier's ranks across all datasets. When comparing 6 algorithms over 85 datasets, \cite{demvsar2006statistical} shows that for the rankings to be significantly different at level $\alpha=0.05$, the critical difference (CD) between the average ranks has to be greater than: 
\begin{equation}
	CD = q_{0.05}(A)\cdot \sqrt{\frac{A(A+1)}{6\cdot N_d}} =  2.850\cdot\sqrt{\frac{42}{510}} \approx  0.82
\end{equation}
The average ranks and critical difference are presented in Figure~\ref{fig:CriticalDiffDiagram}; the critical difference of 0.82 is displayed by the black line.
\modif{It can be seen that COTE ranks highest (average rank of 
2.28),  which is to be expected considering it incorporates the other state-of-the-art algorithms. However, COTE is not ranked significantly higher than Proximity Forest (average rank of 2.88) or ST (average rank of 3.08). 
Proximity Forest is ranked second. Its rank is not significantly different to COTE, ST or BOSS, but it is ranked significantly higher than both EE and DTW.} This affirms Proximity Forest as a classifier with accuracy competitive with the state-of-the-art.

Proximity Forest is the most accurate classifier for 22 of the 85 datasets. However, there is no obvious commonality between these datasets to suggest conditions under which the algorithm is likely to excel. \modif{The detailed accuracy results for Proximity Forest and the five state-of-the-art algorithms are shown in Appendix~\ref{ucr_results}.}

\modif{\subsection{Parameters of Proximity Forest}}
\label{subsec:pf_variations}
\modif{Proximity Forest has two main parameters that merit further investigation. We first explore the sensitivity of accuracy to the number of trees in each ensemble. Then, we discuss the influence of the number of candidates assessed at each node.
A third design choice, random selection of similarity measure per tree as opposed to per node, is explored in Appendix~\ref{ann:on_tree}.}

\subsubsection{On the choice of the number of trees}
\begin{figure}
\centering
\includegraphics[width=.95\linewidth]{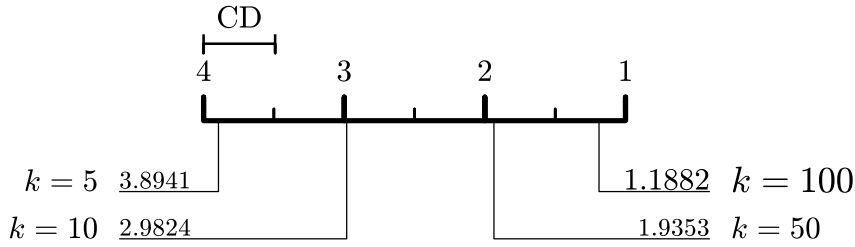}
\caption{Critical difference diagram for Proximity Forest with 5, 10, 50 or 100 trees.}
\label{fig:CD-n-Trees}
\end{figure}

The number of trees is the \modif{first} parameter of the Proximity Forest algorithm with the optimal value being large enough to provide competitive accuracy, yet small enough not to create excessive computational expense. The UCR datasets experiments outlined above were repeated with values of 5, 10, 50, and 100 trees to analyse how many trees were required to meet our needs. \modif{Here, the number of candidates $r$ has been fixed to 1. The Proximity Forest results are averaged over 50 runs.}
Figure~\ref{fig:CD-n-Trees} presents the critical difference diagram for accuracy and different number of trees. As expected, the more trees the higher the average accuracy: models with 100 trees had an average rank of 
1.19 compared to 
1.93, 2.98 and 3.89 for models with 50, 10, and 5 trees respectively. The difference between the highest ranked models 
are large enough to say that models with 100 trees are significantly better than models with 50 trees at the level of alpha equals 0.05.

\begin{figure}
\centering
\includegraphics[width=.6\linewidth]{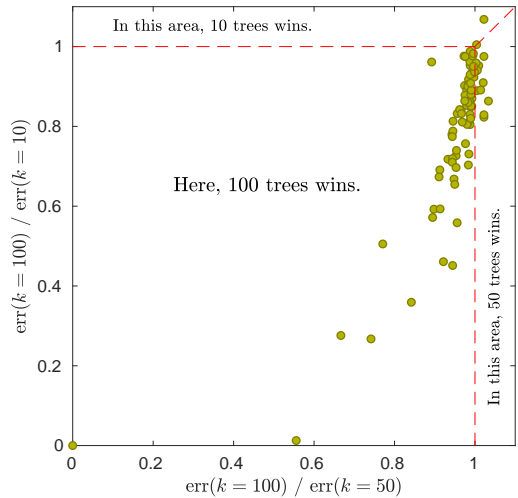}
\caption{Ratio of the error rates of Proximity Forest models: 100 trees over 10 trees (x-axis) against 100 trees over 50 trees (y-axis). A value of less than 1 on either axis indicates that the model with 100 trees has higher accuracy.}
\label{fig:NoTreesPlot}
\end{figure}

Figure~\ref{fig:NoTreesPlot} compares the classification accuracy for 100 trees against 10 and 50 trees by representing them as a ratio of their error rates. Each point represents a single dataset. This shows that having 100 trees is better on most datasets. Moreover, the fact that the data is gathered close to the line with equation $x=1$ shows that it is unlikely that more trees would provide a very significant improvement, because the ratio of error-rates between 100 and 50 is already close to 1 (ie the errors are only slightly reduced). 
We have not experimented with forests comprising more than 100 trees as we felt the computational demands outweighed the expected benefits for our large set of experiments. Memory, training time and testing time all scale linearly with the number of trees, which means that 
doubling the number of trees doubles the required memory and time.
However, 
where computational resources are not an issue, the take home message is that the more trees the better. 

\begin{figure}
\centering
\includegraphics[width=.95\linewidth]{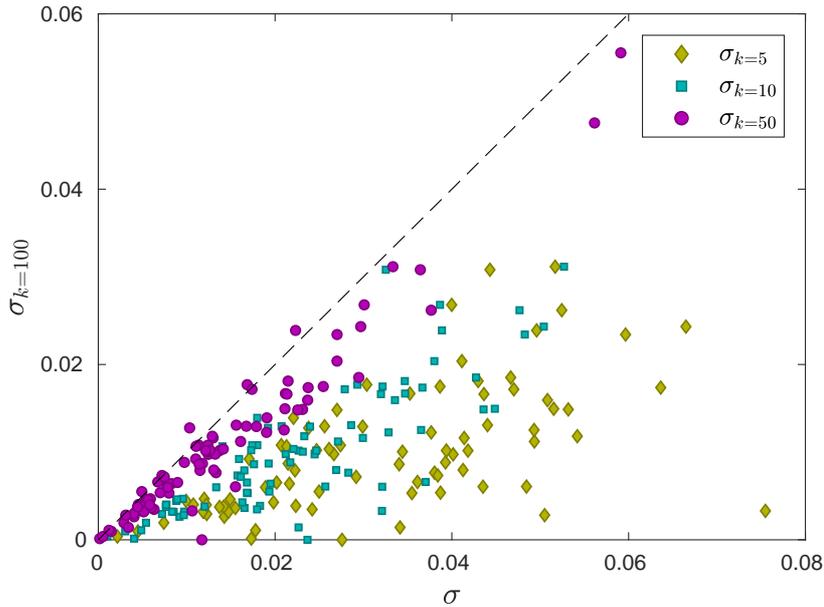}
\caption{Standard deviations $\sigma$ of error rates on the 85 datasets of the UCR archive for Proximity Forest models: 100 trees against 50, 10 and 5 trees.}
\label{fig:stddev}
\end{figure}

As a randomized algorithm, it is finally interesting to study the standard deviation of the errors for Proximity Forest and how it varies with the number of trees. This is what we present in Figure~\ref{fig:stddev} where the y-axis represents the standard deviation on error-rate for 100 trees as a function of the standard deviation on $k$ equals to 5, 10, and 50 trees. Each point represents a single dataset. One can see that the standard deviation reduces as we increase the number of trees, and that the magnitude of this improvement reduces when increasing $k$. Results for 50 trees are starting to be relatively close to the $y=x$ line, showing that only marginal improvements could be expected when going to $k>100$. 

\modif{\subsubsection{Split selection using the Gini index}
\label{subsubsec:candidates_eval}
	This section explores the influence of the number of candidates $r$ that are randomly selected at each node. As a reminder, a set of $r$ candidates -- exemplars and parametrized distance measures -- are evaluated at each node based on the Gini index. The one maximizing the Gini index is retained. To evaluate the influence of $r$, the UCR experiments were repeated for 1, 2, and 5 candidates on 100 trees. The results are averaged over 10 runs.
    
Figure~\ref{fig:gini_influence} compares the classification accuracy for 5 candidates against 1 and 2 candidates. Each point represents the ratio of the error for 5 candidates to that for the alternative on an UCR dataset. %
\begin{figure}
\centering
\includegraphics[width=.6\linewidth]{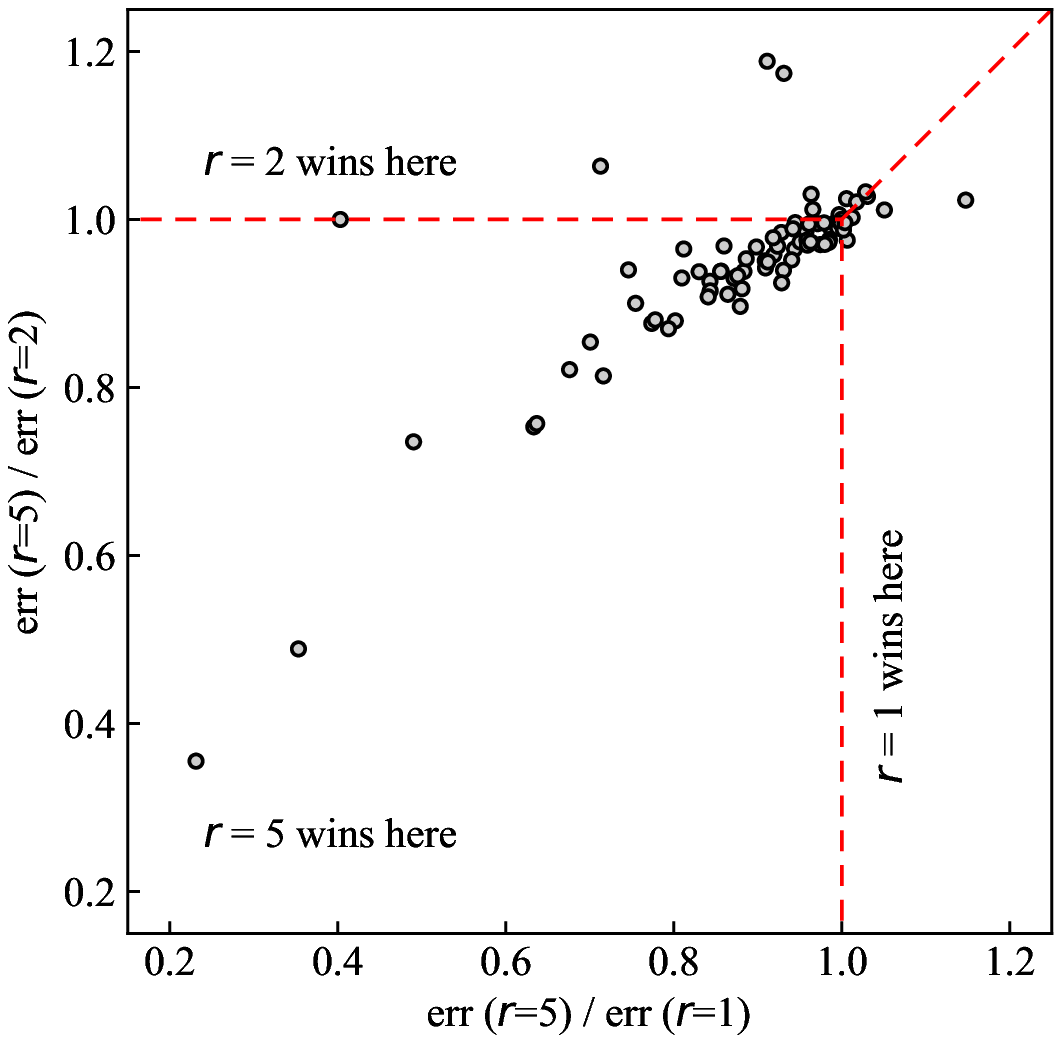}
\caption{Ratio of the error rates of Proximity Forest models: 5 candidates over 1 candidate (x-axis) against 5 candidates over 2 candidates (y-axis). A value of less than 1 on either axis suggests that the model with 5 candidates has superior accuracy}
\label{fig:gini_influence}
\end{figure}
Choosing between 5 candidates results in higher accuracy for most datasets. More precisely, selecting between 5 candidates results in greater accuracy than either 1 or 2 candidates on 61 datasets. Increasing the number of candidates lead to a reduction of the randomness on each node by discarding the worse splitters. Accordingly, the overall Proximity Forest accuracies are improved.
}

\begin{figure}
\centering
\includegraphics[width=.65\linewidth]{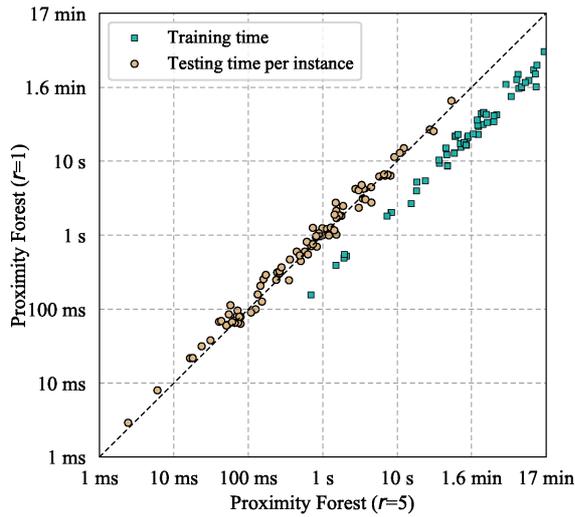}
\caption{Training and testing time of Proximity Forest for 1 and 5 candidates on UCR datasets.}
\label{fig:PF_c1_c5_time}
\end{figure}

\modif{Increasing the number of candidates to more than 5 may further improve the classification accuracy. However, increasing the number of candidates per node has substantial impact on training time. Indeed, the analysis of the Proximity Forest's computational complexity in section~\ref{subsec:computation_analysis} shows that the training time scales linearly with the number of candidates. To verify this analysis, we compare both training and testing time of Proximity Forest for 1 and 5 candidates in Figure~\ref{fig:PF_c1_c5_time}. The testing time is displayed per query. Each point represents a dataset.}
%
%
%
\modif{The results show a mean increase of 4.6 times in training time between 1~and 5 candidates, and a mean decrease of 0.93 times in testing time. 

It is notable that selection between multiple alternatives both reduces testing time and increases the training time by slightly less than the expected multiple of 5 times. This is because it results in slightly shallower trees. Selection of better splits better separates the classes, requiring fewer splits to obtain pure nodes that are made into leaves.}

\modif{The tuning of the number of candidates is therefore driven by a trade-off between accuracy and time.}

\section{Conclusion}
\label{sec:Conclusion}
We introduced Proximity Forest: a novel, scalable algorithm for accurate time series classification. Motivated by a need for an algorithm that could learn from millions of time series, Proximity Forest is an ensemble of trees with a novel splitting criterion that makes it possible to make the most of decades of work in designing time series measures. In our case study, we demonstrated that Proximity Forest 
scales quasi-linearly with the quantity of training data, whereas most state-of-the-art algorithms scale quadratically. 
\modiff{Our experiments on the UCR datasets
show that Proximity Forest is not only very fast. It also has highly competitive accuracy relative to the current state-of-the-art, and is significantly more accurate than EE.}

We believe that there are a number of improvements that can be explored to increase the accuracy of Proximity Forest while maintaining its quasi-linear complexity\modif{, such as improving the randomized selection of parameters for the distance measures -- the current strategy was designed primarily to emulate EE as directly as possible}. We would also like to investigate to what extent this novel algorithm might shed new light on the task of time series indexing. 

\section*{Supplementary material}
To ensure reproducibility, we make available the results of the experiments as well as our source code at \ourRepo{}.

\section*{Acknowledgements}
This research was supported by the Australian Research Council under grant DE170100037.
This material is based upon work supported by the Air Force Office of Scientific Research, Asian Office of Aerospace Research and Development (AOARD) under award number FA2386-17-1-4036. \modiff{We are grateful to the editor and anonymous reviewers whose suggestions and comments have greatly strengthened the paper.}

\newpage 

\appendix
\section{Detailed UCR results}
\label{ucr_results}
\begin{table*}
\centering

\caption{Detailed UCR results for five state-of-the-art algorithms and Proximity Forest. Bold values indicate the best accuracy scores. Proximity Forest results are obtained for 100 trees and 5 candidates, and averaged over 10 runs.}
\tiny
\begin{tabular}{*{11}{l}}
\hline
\textbf{} & \textbf{Train} & \textbf{Test} & \textbf{L} & \textbf{C} & \textbf{DTW} & \textbf{BOSS} & \textbf{ST} & \textbf{EE} & \textbf{COTE} & \textbf{PF} \\
\hline
\textbf{Adiac} & 390 & 391 & 176 & 37 & 60.87 & 76.47 & 78.26 & 66.5 & \textbf{79.03} & 73.4 \\ 
\textbf{ArrHead} & 36 & 175 & 251 & 3 & 80.0 & 83.43 & 73.71 & 81.14 & 81.14 & \textbf{87.54} \\ 
\textbf{Beef} & 30 & 30 & 470 & 5 & 66.67 & 80.0 & \textbf{90.0} & 63.33 & 86.67 & 72.0 \\ 
\textbf{BeetFly} & 20 & 20 & 512 & 2 & 65.0 & \textbf{90.0} & \textbf{90.0} & 75.0 & 80.0 & 87.5 \\ 
\textbf{BirdChi} & 20 & 20 & 512 & 2 & 70.0 & \textbf{95.0} & 80.0 & 80.0 & 90.0 & 86.5 \\ 
\textbf{Car} & 60 & 60 & 577 & 4 & 76.67 & 83.33 & \textbf{91.67} & 83.33 & 90.0 & 84.67 \\ 
\textbf{CBF} & 30 & 900 & 128 & 3 & 99.44 & \textbf{99.78} & 97.44 & \textbf{99.78} & 99.56 & 99.33 \\ 
\textbf{ChloCon} & 467 & 3840 & 166 & 3 & 65.0 & 66.09 & 69.97 & 65.62 & \textbf{72.71} & 63.39 \\ 
\textbf{CinCECG} & 40 & 1380 & 1639 & 4 & 93.04 & 88.7 & 95.43 & 94.2 & \textbf{99.49} & 93.43 \\ 
\textbf{Coffee} & 28 & 28 & 286 & 2 & \textbf{100.0} & \textbf{100.0} & 96.43 & \textbf{100.0} & \textbf{100.0} & \textbf{100.0} \\ 
\textbf{Comput} & 250 & 250 & 720 & 2 & 62.4 & \textbf{75.6} & 73.6 & 70.8 & 74.0 & 64.44 \\ 
\textbf{CricketX} & 390 & 390 & 300 & 12 & 77.95 & 73.59 & 77.18 & \textbf{81.28} & 80.77 & 80.21 \\ 
\textbf{CricketY} & 390 & 390 & 300 & 12 & 75.64 & 75.38 & 77.95 & 80.51 & \textbf{82.56} & 79.38 \\ 
\textbf{CricketZ} & 390 & 390 & 300 & 12 & 73.59 & 74.62 & 78.72 & 78.21 & \textbf{81.54} & 80.1 \\ 
\textbf{DiaSizRed} & 16 & 306 & 345 & 4 & 93.46 & 93.14 & 92.48 & 94.44 & 92.81 & \textbf{96.57} \\ 
\textbf{DisPhAG} & 400 & 139 & 80 & 3 & 62.59 & 74.82 & \textbf{76.98} & 69.06 & 74.82 & 73.09 \\ 
\textbf{DisPhOC} & 600 & 276 & 80 & 2 & 72.46 & 72.83 & 77.54 & 72.83 & 76.09 & \textbf{79.28} \\ 
\textbf{DisPhTW} & 400 & 139 & 80 & 6 & 63.31 & 67.63 & 66.19 & 64.75 & \textbf{69.78} & 65.97 \\ 
\textbf{Earthqua} & 322 & 139 & 512 & 2 & 72.66 & 74.82 & 74.1 & 74.1 & 74.82 & \textbf{75.4} \\ 
\textbf{ECG200} & 100 & 100 & 96 & 2 & 88.0 & 87.0 & 83.0 & 88.0 & 88.0 & \textbf{90.9} \\ 
\textbf{ECG5000} & 500 & 4500 & 140 & 5 & 92.51 & 94.13 & 94.38 & 93.87 & \textbf{94.6} & 93.65 \\ 
\textbf{ECG5days} & 23 & 861 & 136 & 2 & 79.67 & \textbf{100.0} & 98.37 & 82.0 & 99.88 & 84.92 \\ 
\textbf{ElecDev} & 8926 & 7711 & 96 & 7 & 63.08 & \textbf{79.92} & 74.7 & 66.29 & 71.33 & 70.6 \\ 
\textbf{FaceAll} & 560 & 1690 & 131 & 14 & 80.77 & 78.17 & 77.87 & 84.85 & \textbf{91.78} & 89.38 \\ 
\textbf{FaceFour} & 24 & 88 & 350 & 4 & 89.77 & \textbf{100.0} & 85.23 & 90.91 & 89.77 & 97.39 \\ 
\textbf{FacesUCR} & 200 & 2050 & 131 & 14 & 90.78 & \textbf{95.71} & 90.59 & 94.49 & 94.24 & 94.59 \\ 
\textbf{50Words} & 450 & 455 & 270 & 50 & 76.48 & 70.55 & 70.55 & 81.98 & 79.78 & \textbf{83.14} \\ 
\textbf{Fish} & 175 & 175 & 463 & 7 & 83.43 & \textbf{98.86} & \textbf{98.86} & 96.57 & 98.29 & 93.49 \\ 
\textbf{FordA} & 3601 & 1320 & 500 & 2 & 66.52 & 92.95 & \textbf{97.12} & 73.79 & 95.68 & 85.46 \\ 
\textbf{FordB} & 3636 & 810 & 500 & 2 & 59.88 & 71.11 & \textbf{80.74} & 66.17 & 80.37 & 71.49 \\ 
\textbf{GunPoint} & 50 & 150 & 150 & 2 & 91.33 & \textbf{100.0} & \textbf{100.0} & 99.33 & \textbf{100.0} & 99.73 \\ 
\textbf{Ham} & 109 & 105 & 431 & 2 & 60.0 & 66.67 & \textbf{68.57} & 57.14 & 64.76 & 66.0 \\ 
\textbf{HandOut} & 1000 & 370 & 2709 & 2 & 87.84 & 90.27 & \textbf{93.24} & 88.92 & 91.89 & 92.14 \\ 
\textbf{Haptics} & 155 & 308 & 1092 & 5 & 41.56 & 46.1 & \textbf{52.27} & 39.29 & \textbf{52.27} & 44.45 \\ 
\textbf{Herring} & 64 & 64 & 512 & 2 & 53.12 & 54.69 & \textbf{67.19} & 57.81 & 62.5 & 57.97 \\ 
\textbf{InlSkate} & 100 & 550 & 1882 & 7 & 38.73 & 51.64 & 37.27 & 46.0 & 49.45 & \textbf{54.18} \\ 
\textbf{InsWinSou} & 220 & 1980 & 256 & 11 & 57.37 & 52.32 & 62.68 & 59.49 & \textbf{65.25} & 61.87 \\ 
\textbf{ItPowDem} & 67 & 1029 & 24 & 2 & 95.53 & 90.86 & 94.75 & 96.21 & 96.11 & \textbf{96.71} \\ 
\textbf{LaKitAp} & 375 & 375 & 720 & 3 & 79.47 & 76.53 & \textbf{85.87} & 81.07 & 84.53 & 78.19 \\ 
\textbf{Light2} & 60 & 61 & 637 & 2 & 86.89 & 83.61 & 73.77 & \textbf{88.52} & 86.89 & 86.56 \\ 
\textbf{Light7} & 70 & 73 & 319 & 7 & 71.23 & 68.49 & 72.6 & 76.71 & 80.82 & \textbf{82.19} \\ 
\textbf{Mallat} & 55 & 2345 & 1024 & 8 & 91.43 & 93.82 & \textbf{96.42} & 93.99 & 95.39 & 95.76 \\ 
\textbf{Meat} & 60 & 60 & 448 & 3 & \textbf{93.33} & 90.0 & 85.0 & \textbf{93.33} & 91.67 & \textbf{93.33} \\ 
\textbf{MedImg} & 381 & 760 & 99 & 10 & 74.74 & 71.84 & 66.97 & 74.21 & 75.79 & \textbf{75.82} \\ 
\textbf{MidPhAG} & 400 & 154 & 80 & 3 & 51.95 & 54.55 & \textbf{64.29} & 55.84 & 63.64 & 56.23 \\ 
\textbf{MidPhOC} & 600 & 291 & 80 & 2 & 76.63 & 78.01 & 79.38 & 78.35 & 80.41 & \textbf{83.64} \\ 
\textbf{MidPhTW} & 399 & 154 & 80 & 6 & 50.65 & 54.55 & 51.95 & 51.3 & \textbf{57.14} & 52.92 \\ 
\textbf{MotStr} & 20 & 1252 & 84 & 2 & 86.58 & 87.86 & 89.7 & 88.26 & \textbf{93.69} & 90.24 \\ 
\textbf{NoECGT1} & 1800 & 1965 & 750 & 42 & 82.9 & 83.82 & \textbf{94.96} & 84.58 & 93.13 & 90.66 \\ 
\textbf{NoECGT2} & 1800 & 1965 & 750 & 42 & 87.02 & 90.08 & \textbf{95.11} & 91.35 & 94.55 & 93.99 \\ 
\textbf{OliveOil} & 30 & 30 & 570 & 4 & 86.67 & 86.67 & \textbf{90.0} & 86.67 & \textbf{90.0} & 86.67 \\ 
\textbf{OSULeaf} & 200 & 242 & 427 & 6 & 59.92 & 95.45 & \textbf{96.69} & 80.58 & \textbf{96.69} & 82.73 \\ 
\textbf{PhalOC} & 1800 & 858 & 80 & 2 & 76.11 & 77.16 & 76.34 & 77.27 & 77.04 & \textbf{82.35} \\ 
\textbf{Phoneme} & 214 & 1896 & 1024 & 39 & 22.68 & 26.48 & 32.07 & 30.49 & \textbf{34.92} & 32.01 \\ 
\textbf{Plane} & 105 & 105 & 144 & 7 & \textbf{100.0} & \textbf{100.0} & \textbf{100.0} & \textbf{100.0} & \textbf{100.0} & \textbf{100.0} \\ 
\textbf{ProPhAG} & 400 & 205 & 80 & 3 & 78.54 & 83.41 & 84.39 & 80.49 & \textbf{85.37} & 84.63 \\ 
\textbf{ProPhOC} & 600 & 291 & 80 & 2 & 79.04 & 84.88 & \textbf{88.32} & 80.76 & 86.94 & 87.32 \\ 
\textbf{ProPhTW} & 400 & 205 & 80 & 6 & 76.1 & 80.0 & \textbf{80.49} & 76.59 & 78.05 & 77.9 \\ 
\textbf{RefrigDev} & 375 & 375 & 720 & 3 & 44.0 & 49.87 & \textbf{58.13} & 43.73 & 54.67 & 53.23 \\ 
\textbf{ScrType} & 375 & 375 & 720 & 3 & 41.07 & 46.4 & 52.0 & 44.53 & \textbf{54.67} & 45.52 \\ 
\textbf{ShapSim} & 20 & 180 & 500 & 2 & 69.44 & \textbf{100.0} & 95.56 & 81.67 & 96.11 & 77.61 \\ 
\textbf{ShapAll} & 600 & 600 & 512 & 60 & 80.17 & \textbf{90.83} & 84.17 & 86.67 & 89.17 & 88.58 \\ 
\textbf{SmKitAp} & 375 & 375 & 720 & 3 & 67.2 & 72.53 & \textbf{79.2} & 69.6 & 77.6 & 74.43 \\ 
\textbf{SonyAIR1} & 20 & 601 & 70 & 2 & 69.55 & 63.23 & 84.36 & 70.38 & 84.53 & \textbf{84.58} \\ 
\textbf{SonyAIR2} & 27 & 953 & 65 & 2 & 85.94 & 85.94 & 93.39 & 87.83 & \textbf{95.17} & 89.63 \\ 
\textbf{StarCur} & 1000 & 8236 & 1024 & 3 & 89.83 & 97.78 & 97.85 & 92.61 & 97.96 & \textbf{98.13} \\ 
\textbf{Strawber} & 613 & 370 & 235 & 2 & 94.59 & \textbf{97.57} & 96.22 & 94.59 & 95.14 & 96.84 \\ 
\textbf{SwedLeaf} & 500 & 625 & 128 & 15 & 84.64 & 92.16 & 92.8 & 91.52 & \textbf{95.52} & 94.66 \\ 
\textbf{Symbols} & 25 & 995 & 398 & 6 & 93.77 & \textbf{96.68} & 88.24 & 95.98 & 96.38 & 96.16 \\ 
\textbf{SynCon} & 300 & 300 & 60 & 6 & 98.33 & 96.67 & 98.33 & 99.0 & \textbf{100.0} & 99.53 \\ 
\textbf{ToeSeg1} & 40 & 228 & 277 & 2 & 75.0 & 93.86 & 96.49 & 82.89 & \textbf{97.37} & 92.46 \\ 
\textbf{ToeSeg2} & 36 & 130 & 343 & 2 & 90.77 & \textbf{96.15} & 90.77 & 89.23 & 91.54 & 86.23 \\ 
\textbf{Trace} & 100 & 100 & 275 & 4 & 99.0 & \textbf{100.0} & \textbf{100.0} & 99.0 & \textbf{100.0} & \textbf{100.0} \\ 
\textbf{2LeECG} & 23 & 1139 & 82 & 2 & 86.83 & 98.07 & \textbf{99.74} & 97.1 & 99.3 & 98.86 \\ 
\textbf{2Patterns} & 1000 & 4000 & 128 & 4 & 99.85 & 99.3 & 95.5 & \textbf{100.0} & \textbf{100.0} & 99.96 \\ 
\textbf{UWaAll} & 1000 & 6164 & 152 & 8 & 96.23 & 93.89 & 94.22 & 96.85 & 96.43 & \textbf{97.23} \\ 
\textbf{UWaX} & 896 & 3582 & 315 & 8 & 77.44 & 76.21 & 80.29 & 80.54 & 82.19 & \textbf{82.86} \\ 
\textbf{UWaY} & 896 & 3582 & 315 & 8 & 70.18 & 68.51 & 73.03 & 72.56 & 75.85 & \textbf{76.15} \\ 
\textbf{UWaZ} & 896 & 3582 & 945 & 8 & 67.5 & 69.49 & 74.85 & 72.36 & 75.04 & \textbf{76.4} \\ 
\textbf{Wafer} & 896 & 3582 & 315 & 2 & 99.59 & 99.48 & \textbf{100.0} & 99.74 & 99.98 & 99.55 \\ 
\textbf{Wine} & 57 & 54 & 234 & 2 & 61.11 & 74.07 & \textbf{79.63} & 57.41 & 64.81 & 56.85 \\ 
\textbf{WordSyn} & 267 & 638 & 270 & 25 & 74.92 & 63.79 & 57.05 & \textbf{77.9} & 75.71 & 77.87 \\ 
\textbf{Worms} & 181 & 77 & 900 & 5 & 53.25 & 55.84 & \textbf{74.03} & 66.23 & 62.34 & 71.82 \\ 
\textbf{Worms2} & 181 & 77 & 900 & 2 & 58.44 & \textbf{83.12} & \textbf{83.12} & 68.83 & 80.52 & 78.44 \\ 
\textbf{Yoga} & 300 & 3000 & 426 & 2 & 84.3 & \textbf{91.83} & 81.77 & 87.9 & 87.67 & 87.86 \\ 
\hline
\textbf{Av. rank} & & & & & 5.18 & 3.65 & 3.08 & 3.95 & \textbf{2.28} & 2.88 \\
\textbf{Wins} & & & &  & 3 & 20 & \textbf{30} & 8 & 26 & 22\\ 
\hline
\end{tabular}
\end{table*}

\newpage
\normalsize
\section{On a variation of the Proximity Forest}
\label{ann:on_tree}
\begin{figure}[!h]
\centering
\includegraphics[width=.6\linewidth]{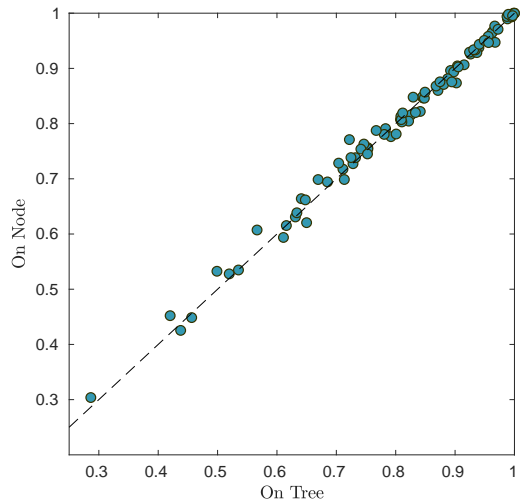}
\caption{Accuracy of Proximity Forest when randomly selecting the distance measure `on node' and `on tree'.}
\label{fig:randomontree}
\end{figure}

\modif{
We decided to explore another variant of the Proximity Forest algorithm by randomly selecting a distance measure for each tree, rather than for each node. In this new variant, only the exemplars and the parameters of the distance-metric are randomly chosen at each node. The UCR experiments were repeated for 100 trees and 1 candidate for this new `on tree' variant. Each Proximity Forest result is averaged over 50 runs.

Figure~\ref{fig:randomontree} compares classification accuracy for the original version `on node', presented in section~\ref{learnPF}, and the proposed variant `on tree'. Each point represents a single dataset of the UCR dataset. The number of trees has been fixed to 100.
}

\modiff{The results show a slight advantage for the `on node' approach with 44 wins, 39 losses and 2 ties. Where the `on tree' variant uses a single distance measure per tree, the `on node' variant allows multiple combinations of measures in a single tree, thus making it more robust to inefficient metrics.}

\end{document}